\documentclass[journal]{IEEEtran}

\usepackage{kotex}
\usepackage{cite}
\usepackage[square, sort,comma,numbers]{natbib}
\usepackage[pdftex]{graphicx}
\usepackage{xcolor}
\usepackage{tikz}
\usepackage{mathtools}
\usepackage{amsmath}
\usepackage{amsfonts}
\usepackage{bm}

\usepackage{booktabs}
\usepackage{subcaption}
\usepackage{array}
\usepackage{multirow}
\usepackage{url}
\usepackage{hyperref}
\usepackage{algorithmicx}
\usepackage[ruled]{algorithm}
\usepackage{algpseudocode}

\newcommand{\eg}{{\em e.g.}}           % e.g.
\newcommand{\ie}{{\em i.e.}}           % i.e.
\newcommand{\bM}{{\bf M}}

\newcommand{\bx}{{\bf x}}
\newcommand{\bX}{{\bf X}}

\newcommand{\bz}{{\bf z}}

\DeclarePairedDelimiterX{\infdivx}[2]{(}{)}{%
  #1\;\delimsize\|\;#2%
}

\newcommand{\chg}{\color{black}}

\begin{document}
% \title{Uncertainty-Gated Stochastic Sequential Imputation for Clinical Time-series and Mortality Prediction}
\title{Uncertainty-Gated Stochastic Sequential Model \\for EHR Mortality Prediction}

\author{Eunji Jun,~\IEEEmembership{Student Member,~IEEE,} Ahmad Wisnu Mulyadi, Jaehun Choi, and~Heung-Il~Suk,~\IEEEmembership{Member,~IEEE}
\thanks{E. Jun and A. W. Mulyadi are with the Department of Brain and Cognitive Engineering, Korea University, Seoul 02841, Republic of Korea (e-mail: ejjun92@korea.ac.kr, wisnumulyadi@korea.ac.kr). J. Choi is with the Medical Information Research Section, Intelligent Convergence Research Laboratory, Electronics and Telecommunications Research Institute(ETRI), Daejeon, 34129, Republic of Korea (e-mail: jhchoi@etri.re.kr). H.-I. Suk is with the Department of Artificial Intelligence and the Department of Brain and Cognitive Engineering, Korea University, Seoul 02841, Republic of Korea (e-mail: hisuk@korea.ac.kr). (\textit{Corresponding author: Heung-Il Suk})}%
}

% \markboth{IEEE TRANSACTIONS ON NEURAL NETWORKS AND LEARNING SYSTEMS}%
% {Shell \MakeLowercase{\textit{et al.}}: Bare Demo of IEEEtran.cls for IEEE Journals}
\markboth{Under Review}% 
{Shell \MakeLowercase{\textit{et al.}}: Bare Demo of IEEEtran.cls for IEEE Journals}
\maketitle

%%%%%%%%%%%%%%%%%%%%%%%%%%%%%%%%%%%%%%%%%%%%%%
\begin{abstract}
Electronic health records (EHR) are characterized as non-stationary, heterogeneous, noisy, and sparse data; therefore, it is challenging to learn the regularities or patterns inherent within them. In particular, sparseness caused mostly by many missing values has attracted the attention of researchers, who have attempted to find a better use of all available samples for determining the solution of a primary target task through the defining a secondary imputation problem. Methodologically, existing methods, either deterministic or stochastic, have applied different assumptions to impute missing values. However, once the missing values are imputed, most existing methods do not consider the fidelity or confidence of the imputed values in the modeling of downstream tasks. Undoubtedly, an erroneous or improper imputation of missing variables can cause difficulties in modeling as well as a degraded performance. In this study, we present a novel variational recurrent network that (i) estimates the distribution of missing variables (\eg, the mean and variance) allowing to represent uncertainty in the imputed values, (ii) updates hidden states by explicitly applying fidelity based on a variance of the imputed values during a recurrence (\ie, uncertainty propagation over time), and (iii) predicts the possibility of in-hospital mortality. It is noteworthy that our model can conduct these procedures in a single stream and learn all network parameters jointly in an end-to-end manner. We validated the effectiveness of our method using the public datasets of MIMIC-III and PhysioNet challenge 2012 by comparing with and outperforming other state-of-the-art methods for mortality prediction considered in our experiments. In addition, we identified the behavior of the model that well represented the uncertainties for the imputed estimates, which indicated a high correlation between the calculated MAE and the uncertainty.
\end{abstract}


\IEEEpeerreviewmaketitle
%%%%%%%%%%%%%%%%%%%%%%%%%%%%%%%%%%%%%%%%%%%%%%
%%%%%%%%%%%%%%%%%%%%%%%%%%%%%%%%%%%%%%%%%%%%%%
\section{Introduction}
\IEEEPARstart{I}{n} the past decade, there have been growing interests and researches in applying machine learning (ML) to clinical domains, particularly on electronic health records (EHR) data analysis for intensive care units \citep{jagannatha2016structured,choi2016medical}. However, owing to the nature of physiological EHR data, they generally involve a substantial number of missing values because of the lack of collection (\ie, unexpected accidents such as equipment damage) or documentation (\ie, an irregular recording across medical variables and even time) \citep{wells2013strategies}. Such an unfavorable characteristic constrains the use of conventional ML models, which commonly assume fully observed and fixed-sized observations in practice \citep{yadav2018mining}. 

To address this issue, some previous studies have directly modeled observations with missing values, for example, transforming them into a time series of distributions over the possible values \citep{zheng2017resolving}. However, not only does this show a low performance under a high rate of missing data, it also requires separate modeling for different datasets. Meanwhile, numerous imputation methods have also been proposed to fill in missing values. Broadly, existing imputation methods can be categorized into (i) \emph{deterministic} or (ii) \emph{stochastic} approaches, depending on the presence of randomness in the imputation process \citep{kalton1986treatment}. Deterministic imputation methods determine only one possible value for each missing value using the model parameters and/or conditions, thus resulting in a unique imputed value for each observation. This approach ranges from statistical methods (\ie, simple mean \citep{kantardzic2011data}, median imputation \citep{acuna2004treatment}, and ratio imputation) to ML approaches, such as expectation maximization (EM) \citep{garcia2010pattern}, $k$-nearest neighbor (KNN), matrix factorization \citep{koren2009matrix}, and matrix completion \citep{mazumder2010spectral}. In recent years, centered on deep learning (DL), recurrent neural networks (RNNs) such as long short-term memory (LSTM) and a gated recurrent unit (GRU) have shown remarkable a performance in modeling the temporal dependencies of a clinical time series and explicitly estimating the missing values, thus being regarded as de facto methods \citep{che2018recurrent,yoon2017multi}. 

However, stochastic methods, for example, multivariate imputation by chained equations (MICE) \citep{azur2011multiple}, possess some inherent randomness and take into account the distributions, thus allowing the generation of samples. More recently, \citep{luo2018multivariate} exploited the adversarial learning framework \citep{goodfellow2014generative}, in which a generator imputes missing values based on other observed values and a discriminator criticizes whether the completed values (by applying both the observed and imputed values) are realistic. In addition, a variational autoencoder (VAE) \citep{kingma:vae,rezende2014stochastic} has also been used for a time series imputation, where the temporal dynamics is separately modeled through a Gaussian process \citep{fortuin2019multivariate} or an RNN \citep{jun2019}.

Although previous imputation methods have shown a reasonable performance, there is no doubt that an erroneous or improper imputation of missing data can degrade the performance in downstream tasks \citep{kreindler2016effects}. Therefore, this has resulted in the need to take account of the fidelity for the imputed values. That is, imputed values with low and high fidelity should be treated differently during the modeling. To the best of our knowledge, most existing methods exploiting the aforementioned imputation techniques do not consider the fidelity of the imputed values in the downstream tasks \citep{che2018recurrent,yoon2017multi,cao2018brits,luo2018multivariate}.

In this study, we explicitly utilize the uncertainty for imputing missing values represented in terms of variance as the fidelity and propose a novel \emph{uncertainty-gated stochastic sequential model} for a clinical time-series prediction. Inspired by the success of stochastic RNN models that introduce a stochastic gradient variational Bayes (SGVB) approach \citep{rezende2014stochastic,kingma:vae} into an RNN sequence model, we take advantage of its capabilities to capture the underlying sequential structure and temporally generate missing values for a multivariate time-series data imputation. In addition to providing probabilistic imputation estimates, as a result of stochastic inference, we estimate their corresponding uncertainty from a latent space and further propagate it within the GRU cells in a time-series modeling for mortality prediction. Note that, to harness the rich representational power of a latent space, our proposed method implicitly uses the imputed time series for prediction.
The main contributions of this work are as follows:
\begin{itemize}
	\item To the best of our knowledge, our study is the first to use an extended RNN with stochastic units to provide probabilistic imputation estimates with uncertainty.
	\item We propose a novel GRU cell, called GRU-U, that exploits uncertainty-gated attention and further leverages attention weights for a reliable mortality prediction.
	\item We simultaneously conduct a missing value imputation and further prediction task jointly in an end-to-end manner.
	\item We evaluated our model on real-world healthcare datasets and achieved state-of-the-art results for the mortality prediction task, validating the effectiveness of the proposed approach in a clinical setting.
\end{itemize}

\begin{figure*}[h!]
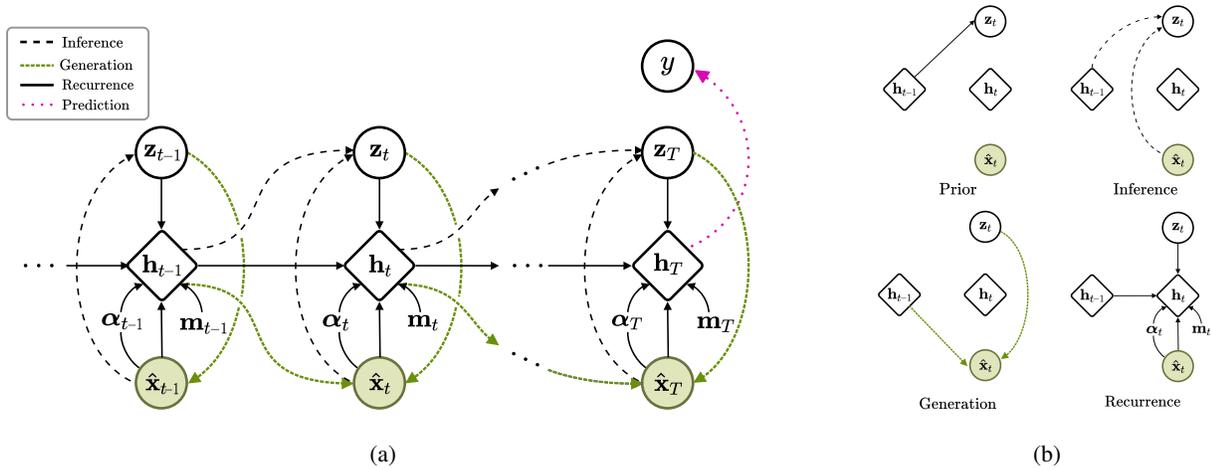

	\centering
	\begin{subfigure}[b]{0.59\linewidth}\centering
		\includegraphics[width=0.95\columnwidth]{./figures/VRNN+Unc.pdf}%\vskip.15in
		\caption{}
% 		\caption{The proposed imputation architecture}
		\label{fig:imputation}
	\end{subfigure}
	\begin{subfigure}[b]{0.37\linewidth}\centering
		\includegraphics[width=0.69\columnwidth]{./figures/VRNN+Unc_timestep.pdf}
		\caption{}
% 		\caption{Each step of the VRNN}
		\label{fig:imputation_t}
	\end{subfigure}
	\caption{Graphical illustrations of (a) the whole architecture of the stochastic recurrent imputation method and (b) each step of the VRNN.}
	\label{fig:overall}
\end{figure*}

\section{Related Work}
In the past decades, numerous imputation methods have been proposed to handle missing values that are sampled irregularly and/or sparsely in a multivariate time series. Conventional imputation methods can largely be divided into three classes. The first comprises statistical imputation methods, from simple mean \citep{kantardzic2011data}, median \citep{acuna2004treatment}, and ratio imputation methods to classical statistical time-series models including the auto-regressive integrated moving average (ARIMA) \citep{ansley1984estimation}, which eliminates the non-stationary parts in a sequence and fits a parameterized stationary model. However, these statistical imputation methods have a limitation in terms of inadequately modeling the temporal sequence and deterministically imputing missing values without any stochastic factors.

The second class comprises a variety of ML-based imputation methods have been developed for better missing value estimation, such as the EM algorithm \citep{garcia2010pattern}, KNN, matrix factorization \citep{koren2009matrix}, and matrix completion \citep{mazumder2010spectral}. Furthermore, MICE \citep{azur2011multiple} is widely used in practice by iteratively applying the aforementioned ML-based methods and averaging the results. Although such methods exploit randomness added as a stochastic approach, they rarely consider the uncertainty information.

Finally, as the thrid class, \ie, deep learning-based imputation methods, the RNNs \citep{che2018recurrent,yoon2017multi} have more recently been proven to achieve successes in modeling temporal dependencies and imputing the sequence, particularly within the healthcare domain. Hence, we focus more on the RNN-based state-of-the-art imputation methods of a clinical time series, which are usually leveraged together with further downstream, \ie, classification/regression.

GRU-D \citep{che2018recurrent} assumes that missing variables can be derived by combining the mean imputation and forward filling using the last observation. For this, a trainable temporal decaying factor toward the global mean is introduced from the time interval information. In addition, the masking vector is directly modeled with the time-series data inside the GRU cell, allowing missing patterns to also be modeled internally. GRU-D achieves a superior performance in various clinical tasks but has a strong assumption regarding the data, which may not be well-suited to typical time-series datasets.
%it assumes that the missing variable is expressed as a combination of last observation and global mean by following properties of healthcare data. 

As another RNN-based imputation method, M-RNN \citep{yoon2017multi} utilizes a bi-directional RNN to reconstruct missing values by operating both within streams (\ie, interpolation) and across streams (\ie, imputation). The imputed values in M-RNN are treated as constants, which cannot be sufficiently updated.

As another bidirectional approach, similar to an M-RNN, BRITS \citep{cao2018brits} also uses bi-directional recurrent dynamics by considering the forward and backward directions on a given time-series to solve the error delay problem until the presence of the next observation. It further formulates a feature-based estimation as a correlation among variables in addition to a history-based estimation. At the same time, it considers the imputed values as variables and updates them during back-propagation as opposed to M-RNN, thereby resulting in a SOTA performance in the healthcare domain. 
% {\todo However, its main drawback is difficulty of practical application in real-time healthcare domain due to the approach of dynamically feeding inputs into the model backward.}

% In the meantime, as the self-attention mechanism \citep{vaswani2017attention} without any recurrence surged in the sequence-to-sequence modeling, \citep{song2018attend} \etal\ employed a masked self-attention mechanism via positional encoding and dense interpolation strategies for clinical time series modeling, and showed reasonable performance in the EHR multi-tasks.

Despite these remarkable realizations of deep learning-based imputation models, it should be noted that these methods have a major drawback in that they do not investigate the uncertainty for the imputation estimates. Meanwhile, \citep{luo2018multivariate} recently proposed the GRU-I that uses the adversarial learning framework to impute a multivariate time-series and explicitly leverage it for further classification. However, this method has a limitation in that the imputation and downstream steps are separated, and thus training dose not occur in an end-to-end manner. In addition, although an uncertainty occurs in an input space, it is not utilized explicitly for classification tasks.

By adding stochasticity into the hidden states of an RNN, this study is the first to provide probabilistic imputation estimates with uncertainty, and further apply uncertainty-gated attention within the GRU cell, as well as consider the effectiveness in terms of an in-hospital mortality prediction task.

%%%%%%%%%%%%%%%%%%%%%%%%%%%%%%%%%%%%%%%%%%%%%%
%%%%%%%%%%%%%%%%%%%%%%%%%%%%%%%%%%%%%%%%%%%%%%
\section{Methods}
In this section, we describe our proposed \emph{uncertainty-gated stochastic sequential model} that extends a variational RNN (VRNN) \citep{chung2015recurrent} for mortality prediction using EHR data. In particular, by leveraging a VRNN as our base model, we devise a novel network that handles representation learning, missing value imputation, and in-hospital mortality prediction simultaneously. Notably, we propose a new type of GRU cell, in which temporal information encoding in hidden states is updated by propagating the uncertainty for the inferred distribution over the variables. The overall architecture of the proposed model is shown in Fig. \ref{fig:overall}.

Our proposed method consists of three parts: (i) stochastic recurrent missing value imputation, (ii) uncertainty-gated attention, and (iii) in-hospital mortality prediction. 

%%%%%%%%%%%%%%%%%%%%%%%%%%%%%%%%%%%%%%%%%%%%%%
\subsection{Data Representation}
Given a multivariate time series with $D$ variables over $T$ time points, we denote it as $\bX=(\bx_{1},...,\bx_{t},..., \bx_{T})^\top\in\mathbb{R}^{T\times D}$, where $\bx_{t}\in\mathbb{R}^D$ represents the $t$-th observation of all variables {\chg observed at timestamp $s_t$,} and $x_{t}^d$ is the $d$-th element or variable in $\bx_{t}$. In this setting, because the time-series $\bX$ includes missing values, we introduce the masking vector across the time-series, $\mathbf{M}=(\mathbf{m}_{1},...,\mathbf{m}_{t}, ..., \mathbf{m}_{T})^\top\in\mathbb{R}^{T\times D}$, with the same size of $\bX$, to mark which variables are observed or missing. In particular, $m_{t}^d=1$ if $x_{t}^d$ is observed, $m_{t}^d=0$, otherwise. Considering a masking vector, we define a new multivariate time series including missing values, $\tilde{\bX}=(\tilde{\bx}_{1}, ..., \tilde{\bx}_{t}, ..., \tilde{\bx}_{T})^\top\in\mathbb{R}^{T\times D}$, as follows:
\begin{equation}\label{eq1}
\tilde{x}_{t}^d=
	\begin{cases}
		x_{t}^d, & \text{if $m_{t}^d=1$}\\
        *, & \text{otherwise}\\	
	\end{cases}
\end{equation}	
where * indicates an unobserved value to be estimated by the proposed imputation method. Initially, we set * in $\tilde{\bX}$ to zero \citep{luo2018multivariate, nazabal2018handling}. {\chg In addition, we maintain the \emph{time interval}, $\boldsymbol{\Delta}=(\boldsymbol{\delta}_{1},...,\boldsymbol{\delta}_{t}, ..., \boldsymbol{\delta}_{T})^\top\in\mathbb{R}^{T\times D}$, defined as the difference between the last observation and the current timestamp following the equations for each variable $\delta_{t}^d$:
\begin{equation}\label{time_interval}
\delta_{t}^d=
	\begin{cases}
		s_{t}^d - s_{t-1}^d + \delta_{t-1}^d, & \text{if $t>1$, $m_{t-1}^d=0$}\\
        s_{t}^d - s_{t-1}^d, & \text{if $t>1$, $m_{t-1}^d=1$}\\	
        1, & \text{if $t=1$}
	\end{cases}.
\end{equation}	

}

Given a clinical time series dataset $\mathcal{D}=\{(\tilde{\bX}^{(n)}, \mathbf{M}^{(n)}, \boldsymbol{\Delta}^{(n)})\}_{n=1}^{N}$ for $N$ subjects, we define an in-hospital mortality prediction as a binary classification problem with labels $y^{(n)}\in\{0,1\}$. To avoid cluttering, and without loss of generality, we simply use functional notation $(\tilde{\mathbf{x}}_{t},\mathbf{m}_{t},\boldsymbol{\delta}_{t})$ for a patient's EHR at time $t$, ignoring a superscript $(n)$.

%%%%%%%%%%%%%%%%%%%%%%%%%%%%%%%%%%%%%%%%%%%%%%
\subsection{Stochastic Recurrent Missing Value Imputation}
\label{subsec:Stochastic_Recurrent_Missing_Value_Imputation}
Inspired from a finding indicating that the use of stochastic representations for hidden states in RNNs helps improve the time-series modeling \citep{fabius2014variational,chung2015recurrent}, we adopt a VRNN \citep{chung2015recurrent} in our base model architecture.

The VRNN is a probabilistic extension of an RNN with stochastic units. In particular, the hidden states of RNNs include latent random variables by combining the elements of the VAE. This allows modeling their distributional characteristics in a latent space, where the underlying structure of sequential data can be better represented. Based on the estimated distribution of latent variables, it becomes possible to generate input values, with which we can impute missing values accordingly. 

The overall stochastic imputation process comprises a series of steps as shown in Fig. \ref{fig:overall}: (i) \emph{prior} and (ii) \emph{posterior inference} over the latent variable, (iii) estimating the observational distribution, called a \emph{generation}, and (iv) representing hidden states by \emph{recurrence}. In other words, the VRNN iteratively performs both the inference and generation process at every time step. Note that the inference step approximates the true posterior, and the generation step performs the imputation by reconstructing data from this posterior. To handle output probability distributions and stochastic training, arbitrarily flexible functions such as neural networks can be chosen.

%%%%%%%%%%%%%%%%%%%%%%%%%%%%%%%%%%%%%%%%%%%%%%
\subsubsection{Prior}
The prior distribution on the latent random variable follows the distribution $p(\bz_t;{\theta_{0}})=\mathcal{N}\left(\boldsymbol{\mu}_{0,t}, \text{diag}(\boldsymbol{\sigma}_{0,t}^2)\right)$,
\begin{equation}
	[\boldsymbol{\mu}_{0,t}, \log\boldsymbol{\sigma}_{0,t}^2]=\mathcal{F}^{\text{prior}}(\mathbf{h}_{t-1}),
	\label{eq:prior}
\end{equation}
where $\mathcal{F}^{\text{prior}}$ is a function with learnable parameters $\theta_0$ and previous hidden states $\mathbf{h}_{t-1}$ as input (Prior in Fig. \ref{fig:imputation_t}). In a clinical setting, this latent variable $\bz_t$ can be interpreted as the patient's hidden health status at a particular time point. 

%%%%%%%%%%%%%%%%%%%%%%%%%%%%%%%%%%%%%%%%%%%%%%
\subsubsection{Inference}
During the inference phase (black dashed arrows in Fig. \ref{fig:imputation} and Inference in Fig. \ref{fig:imputation_t}), we aim to learn the inference network that approximates the true posterior distribution over the latent variables $p({\mathbf z}_t|\tilde{{\mathbf x}}_t)$ to $q({\mathbf z}_t|\tilde{\mathbf x}_t;{\phi})=\mathcal{N}\left(\boldsymbol{\mu}_{\bz,t}, \text{diag}(\boldsymbol{\sigma}_{\bz,t}^2)\right)$. We estimate the mean and log-variance using $\mathcal{F}^{\text{inf}}$ with the parameter $\phi$, which is conditioned on both $\tilde{\bx}_t$ and $\mathbf{h}_{t-1}$ as:
\begin{equation}
	[\boldsymbol{\mu}_{\bz,t}, \log\boldsymbol{\sigma}_{\bz,t}^2]=\mathcal{F}^{\text{inf}}(\mathcal{F}^{\bx}(\tilde{\bx}_t),\mathbf{h}_{t-1}), \label{eq:inference}
\end{equation}
where $\mathcal{F}^{\bx}$ is a non-linear feature extractor from $\tilde{\bx}_t$. For all time steps, an approximate posterior depending on $\bz_{1:T}$ and $\tilde{\bx}_{1:T}$ factorizes as follows:
\begin{gather}
	q(\bz_{1:T}|\tilde{\bx}_{1:T})=q(\bz_{1}|\tilde{\bx}_{1})\prod_{t=2}^Tq(\bz_t|\bz_{1:t-1}, \tilde{\bx}_{1:t}).\label{eq:factorize}
\end{gather}
Here, a reparameterization trick \citep{kingma:vae,rezende2014stochastic} is used to make the network differentiable in our implementation, as in an auto-encoding variational Bayes algorithm. We sample $\boldsymbol{\epsilon}\sim\mathcal{N}(\mathbf{0},\mathbf{I})$ and then represent $\bz_t=\boldsymbol{\mu}_{\bz,t}+\boldsymbol{\sigma}_{\bz,t}\odot\boldsymbol{\epsilon}$ with $\boldsymbol{\mu}_{\bz,t}$ and $\boldsymbol{\sigma}_{\bz,t}$ estimated from an inference network in Eq. \eqref{eq:inference}, where $\odot$ denotes an element-wise multiplication. It should be noted that the inference step is implicitly involved in our imputation process in the context of the approximated posteriors' contribution to the ensuing generation step.

%%%%%%%%%%%%%%%%%%%%%%%%%%%%%%%%%%%%%%%%%%%%%%
\subsubsection{Generation}
The generation step (green dashed arrows in Fig. \ref{fig:imputation} and Generation in Fig. \ref{fig:imputation_t}) learns the generative network following the reconstruction distribution of $p(\tilde{\mathbf x}_t|{\mathbf z}_t;{\theta})=\mathcal{N}\left(\boldsymbol{\mu}_{\bx,t}, \text{diag}(\boldsymbol{\sigma}_{\bx,t}^2)\right)$, the mean and log-variance of which are estimated as:
\begin{equation}
	[\boldsymbol{\mu}_{\bx,t}, \log\boldsymbol{\sigma}_{\bx,t}^2]=\mathcal{F}^{\text{gen}}(\mathcal{F}^{\bz}(\bz_t),\mathbf{h}_{t-1}),
	\label{eq:generation}
\end{equation}
where $\mathcal{F}^{\text{gen}}$ is a generating function with parameter $\theta$, and $\mathcal{F}^{\bz}$ is a non-linear feature extractor from $\bz_t$. Note that they are dependent on the posterior ${\bz}_t$ approximated at the inference step, as well as $\mathbf{h}_{t-1}$. The joint distribution across subsequent time steps also factorizes as: 
\begin{align}
	p(\tilde\bx_{1:T}, &\bz_{1:T})=p(\bz_1)p(\tilde{\bx}_1|\bz_1)\nonumber
\\ &\prod_{t=2}^Tp(\bz_t|\tilde{\bx}_{1:t-1},\bz_{1:t-1})p(\tilde{\bx}_t|\bz_{1:t},\tilde{\bx}_{1:t-1})
\end{align}
where $p(\bz_t|\tilde{\bx}_{1:t-1},\bz_{1:t-1})$ and $p(\tilde{\bx}_t|\bz_{1:t},\tilde{\bx}_{1:t-1})$ can be obtained from the prior distribution in Eq. \eqref{eq:prior} and reconstruction distribution in Eq. \eqref{eq:generation} at time $t$, respectively.

{\chg  
The imputation is performed using the mean values $\boldsymbol{\mu}_{\bx,t}$ from the given reconstruction distribution. Here, we consider a temporal decaying factor and a feature correlation of a multivariate time-series. Considering a temporal context where the influence of the medical features fades over time for the case of missing long-term values, it is well suited to introduce a decay mechanism for EHR time series modeling \citep{che2018recurrent}. Following the aforementioned properties, the negative exponential rectifier is exploited to make the temporal decay rate $\boldsymbol{\gamma}_t$ monotonically decrease, which is denoted from the time interval $\boldsymbol{\delta}_t$ as follows:
\begin{equation}
    \boldsymbol{\gamma}_t=\exp\{-\max(\mathbf{0}, \mathbf{W}_{\gamma}\boldsymbol{\delta}_t+\mathbf{b}_{\gamma})\}.
\end{equation}
We learn the decay rates from the training data by learning the model parameters $\mathbf{W}_{\gamma}, \mathbf{b}_{\gamma}$, rather than fixed a priori. The actual observations and the estimated mean values from the VRNN are then combined by the weight $\boldsymbol{\beta}_t\in[0,1]^D$ determined from the temporal decay rate and masking vector as follows:
\begin{gather}
    \boldsymbol{\beta}_t=\sigma\left(\mathbf{W}_{\beta}[\boldsymbol{\gamma}_t \circ \mathbf{m}_t]+\mathbf{b}_{\beta}\right),\\
    \mathbf{c}_t=\boldsymbol{\beta}_t \odot\tilde{\bx}_t+(\mathbf{1}-\boldsymbol{\beta}_t)\odot  \boldsymbol{\mu}_{\bx,t},
\end{gather}
where $\mathbf{W}_{\beta}$ and $\mathbf{b}_{\beta}$ are learnable parameters. This temporal decaying mechanism ensures that missing values are smoothly replaced over time. In addition, we further introduce additional feature correlations into the imputation process, where one feature is represented by a linear combination of the others
\begin{equation}
    \tilde{\mathbf{c}}_t=\mathbf{W}_{x}\boldsymbol{\mu}_{\bx,t}+\mathbf{b}_{x},
\end{equation}
where $\mathbf{W}_{x}$ is a parameter matrix with a diagonal of all zeros.

Consequently, the temporally decayed estimates $\mathbf{c}_t$ and feature-correlated estimates $\tilde{\mathbf{c}}_t$ are fully integrated into a \emph{combination vector} $\hat{\mathbf{c}}_t$ using a $1\times1$ convolution operation ($*$), followed by a max pooling operation that incorporates all channel information as:
\begin{equation}
    \hat{\mathbf{c}}_t=\text{max-pool}\left(\mathbf{c}_t*\tilde{\mathbf{c}}_t\right).
\end{equation}
Thus, depending on the presence of the observation from masking values ${\mathbf{m}}_t$, we can formulate the final imputation estimates $\hat{\mathbf{x}}_t$ as follows:
\begin{equation}
    \hat{\mathbf{x}}_t=\mathbf{m}_t \odot\tilde{\bx}_t+(\mathbf{1}-\mathbf{m}_t)\odot  \hat{\mathbf{c}}_t. \label{eq:x_hat}
\end{equation}

In addition, we also capture the \emph{uncertainty} $\mathbf{u}_t$ of the imputation estimates by using $\boldsymbol{\sigma}_{\bx,t}$ from the generative network. Here, the uncertainties of observed values are set to zero, indicating a full fidelity. It is worth noting that the imputed values are stochastically estimated with corresponding uncertainty, considering the posteriors inferred at the current time step. For the estimation of the imputed values, the uncertainties $\mathbf{u}_t$ are evaluated as follows:
\begin{equation}
	{\mathbf{u}}_t=({\mathbf{1}}-\mathbf{m}_t)\odot{\boldsymbol{\sigma}}_{\bx,t}.\label{eq:u}
\end{equation}
}

%%%%%%%%%%%%%%%%%%%%%%%%%%%%%%%%%%%%%%%%%%%%%%
\subsubsection{Recurrence}
During the recurrence phase (black solid arrows in Fig. \ref{fig:imputation} and Recurrence in Fig. \ref{fig:imputation_t}), the extracted features from $\hat{\mathbf{x}}_t$ and $\bz_t$ and the previous hidden state $\mathbf{h}_{t-1}$ are fed into the hidden state as 
\begin{gather}
	{\mathbf h}_t=\mathcal{H}\left(\mathcal{F}^{\bx}(\hat{\mathbf{x}}_t),\mathcal{F}^{\bz}(\bz_t),\mathbf{h}_{t-1}\right),
\end{gather}
where $\mathcal{H}$ is a transformation function conditioned on both $\hat{\mathbf{x}}_{1:T}$ and $\bz_{1:T}$. For the transformation function, in our study, a GRU cell \citep{cho2014learning} is used. Notably, the imputed values $\hat{\mathbf{x}}_t$ and latent variables $\bz_t$ are tied through the deterministic hidden state $\mathbf{h}_t$, which further updates the inferred distribution over variables during the recurrence.

\begin{figure*}[h!]
	\centering
	\begin{subfigure}[b]{0.35\linewidth}\centering
		\includegraphics[width=0.95\columnwidth]{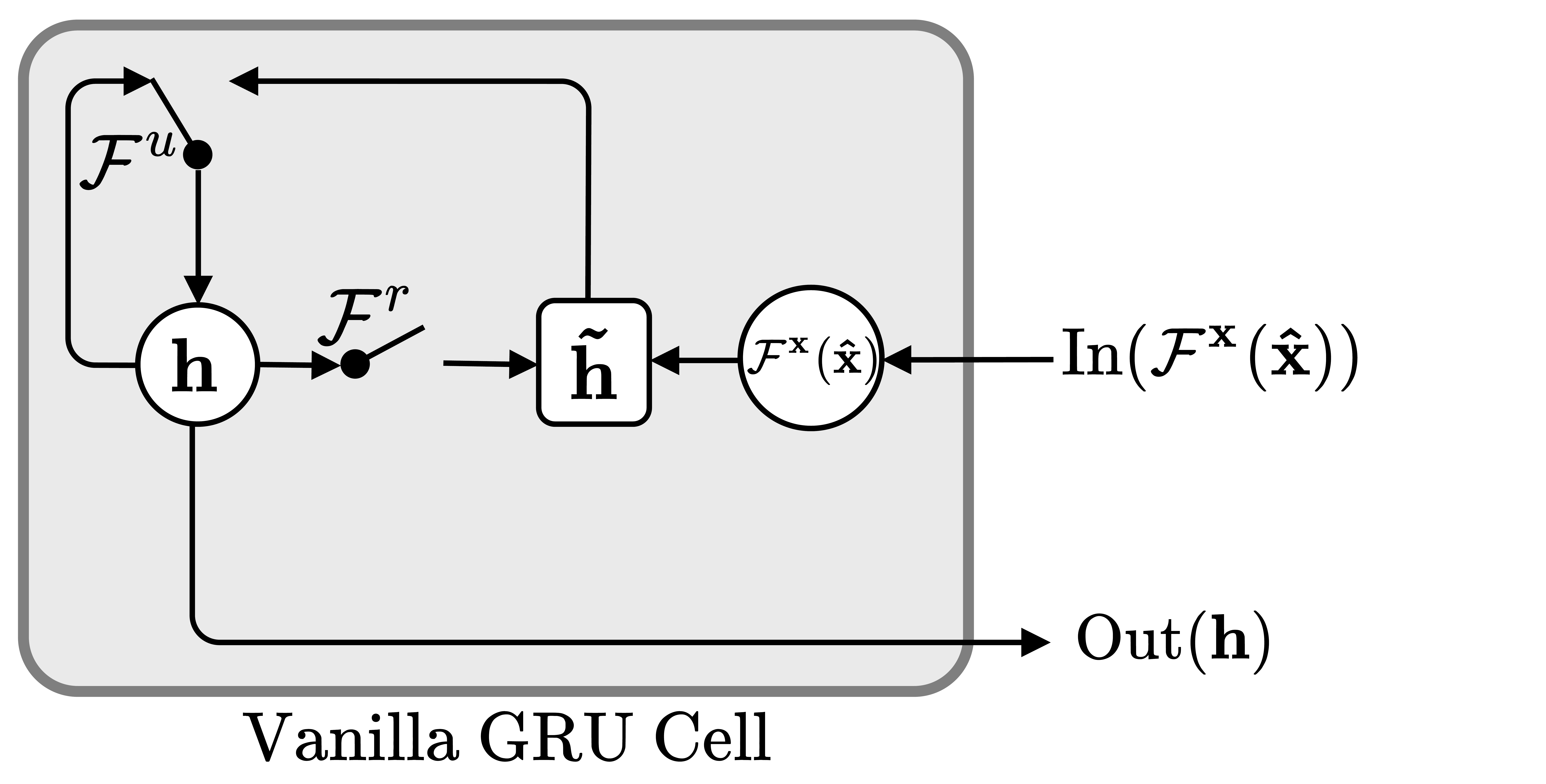}%\vskip.15in
		\caption{}
% 		\caption{Original vanilla GRU cell}
% 		\label{fig:imputation}
	\end{subfigure}
	\hspace{30pt}
% 	\hfill
	\begin{subfigure}[b]{0.35\linewidth}\centering
		\includegraphics[width=0.95\columnwidth]{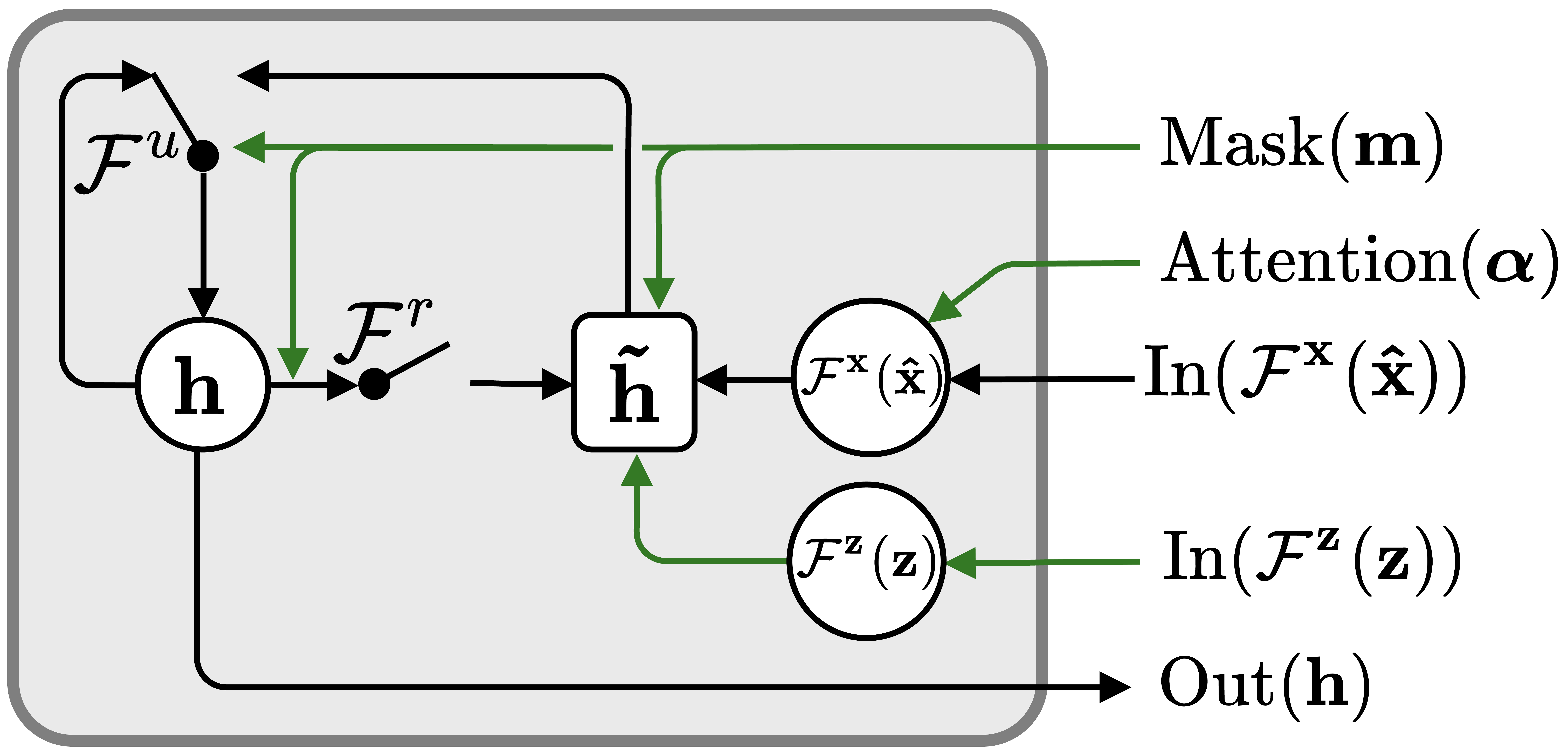}
		\caption{}
% 		\caption{Proposed GRU-U cell}
% 		\label{fig:imputation_t}
	\end{subfigure}
	\caption{Comparison of (a) original vanilla GRU cell and (b) proposed GRU-U cell.}
	\label{fig:gru-u}
\end{figure*}

%%%%%%%%%%%%%%%%%%%%%%%%%%%%%%%%%%%%%%%%%%%%%%
\subsection{Uncertainty-Gated Attention}
\label{subsec:Uncertainty_Gated_Attention}
\begin{figure*}[h]
\begin{gather}
	\mathcal{F}^{r}_t=\varphi\left({\mathbf W}_{r}\left[\mathcal{F}^{\bx}(\hat{\mathbf{x}}_t) \circ \boldsymbol{\alpha}_t \circ \mathcal{F}^{\bz}({\bz}_t) \right]+{\mathbf U}_{r}{\mathbf h}_{t-1}+{\mathbf V}_{r}{\mathbf m}_t+{\mathbf b}_{r}\right)\label{eq:reset_gate}\\
	%\mathcal{F}^{r}_t=\varphi^{r}\left({\mathbf W}_{r}[\mathcal{F}^{\bx}(\hat{\bx}_t)\circ\mathcal{F}^{\bz}({\bz}_t)\circ{\boldsymbol{\alpha}_t}]+{\mathbf U}_{r}{\mathbf h}^{\mathbf u}_{t-1}+{\mathbf V}_{r}{\mathbf m}_t+{\mathbf b}_{r}\right){eq:reset_gate}\\
	\mathcal{F}^{u}_t=\varphi\left({\mathbf W}_{u}\left[\mathcal{F}^{\bx}(\hat{\mathbf{x}}_t) \circ \boldsymbol{\alpha}_t \circ \mathcal{F}^{\bz}({\bz}_t) \right]+{\mathbf U}_{u}{\mathbf h}_{t-1}+{\mathbf V}_{u}{\mathbf m}_t+{\mathbf b}_{u}\right)\label{eq:forget_gate}\\
	%\mathcal{F}^{u}_t=\varphi^{u}\left({\mathbf W}_{u}[\mathcal{F}^{\bx}(\hat{\bx}_t)\circ\mathcal{F}^{\bz}({\bz}_t)\circ{\boldsymbol{\alpha}_t}]+{\mathbf U}_{u}{\mathbf h}^{\mathbf u}_{t-1}+{\mathbf V}_{u}{\mathbf m}_t+{\mathbf b}_{u}\right)\label{eq:forget_gate}\\
	\tilde{\mathbf h}_t=\varphi\left({\mathbf W}_{{h}}\left[\mathcal{F}^{\bx}(\hat{\mathbf{x}}_t)\circ \boldsymbol{\alpha}_t \circ \mathcal{F}^{\bz}({\bz}_t) \right]+{\mathbf U}_{{h}}[\mathcal{F}^{r}_t\odot{\mathbf h}_{t-1}]+{\mathbf V}_{{h}}{\mathbf m}_t+{\mathbf b}_{{h}}\right) \label{eq:tilde_hidden}\\
	%\tilde{\mathbf h}_t^{\mathbf u}=\varphi^{h}\left({\mathbf W}_{{h}}[\mathcal{F}^{\bx}(\hat{\bx}_t)\circ\mathcal{F}^{\bz}({\bz}_t)\circ{\boldsymbol{\alpha}_t}]+{\mathbf U}_{{h}}[\mathcal{F}^{r}_t\odot{\mathbf h}^{\mathbf u}_{t-1}]+{\mathbf V}_{{h}}{\mathbf m}_t+{\mathbf b}_{{h}}\right) \label{eq:tilde_hidden}\\
	{\mathbf h}_t=({\mathbf 1}-\mathcal{F}^{u}_t)\odot{\mathbf h}_{t-1}+\mathcal{F}^{u}_t\odot\tilde{\mathbf h}_t  \label{eq:hidden}
\end{gather}
\end{figure*}
In this study, we investigate the application of uncertainty for a downstream task. For each timestamp, we efficiently propagate the uncertainty within the GRU cell gate such that the imputed data can be axiomatically incorporated with this uncertainty in a non-linear fashion. To this end, the uncertainty ${\bf u}_t$ is additionally injected into the hidden state as input into the GRU cell with the features from $\hat{\mathbf{x}}_t$, $\bz_t$ and the previous hidden state as
\begin{gather}
	{\mathbf h}_t=\mathcal{H}\left(\mathcal{F}^{\bx}(\hat{\mathbf{x}}_t),\mathcal{F}^{\bz}(\bz_t),\mathbf{h}_{t-1}, \mathbf{u}_t\right).
\end{gather}

The basic idea of the proposed approach is to reduce the influence of the imputed data with low fidelity on the subsequent downstream tasks. To this end, our approach introduces a trainable decay designed for uncertainty in the model, and it efficiently propagates through the GRU gate units, which we call \emph{uncertainty-gated attention}. Therefore, in this paper, we propose a novel GRU cell, called GRU-U that further injects uncertainty-gated attention into a vanilla GRU cell. The difference between the vanilla GRU cell and the proposed GRU-U is schematically depicted in Fig. \ref{fig:gru-u}. Similar to the temporal decaying rate, the attention weights $\boldsymbol{\alpha}_t\in(\mathbf{0},\mathbf{1}]$ are formally:
\begin{equation}
	\boldsymbol{\alpha}_t=\exp\{-\max(\mathbf{0},{\mathbf W}_{\alpha}{\mathbf{u}_t}+{\mathbf b}_{\alpha})\},\label{eq:attention}
\end{equation}
where ${\mathbf W}_{\alpha}$ and ${\mathbf b}_{\alpha}$ are model parameters that are jointly learned using other recurrence parameters. In particular, ${\mathbf W}_{\alpha}$ is restricted to a diagonal matrix to ensure that the decay factor of each variable is independent of the others. 

% \begin{figure}[h]
% \centering
% {\includegraphics[width=0.8\columnwidth]{./img/GRU-U.pdf}}
% \caption{The vanilla GRU cell and the proposed GRU-U.}\label{fig:gru-u}
% \end{figure}
  
In GRU-U, there are two types of gates: a \emph{reset gate} $\mathcal{F}^{r}$ and an \emph{update gate} $\mathcal{F}^{u}$ used to control the information. Compared with the vanilla GRU cell, the attention weight vector $\boldsymbol{\alpha}_t$ and masking vector ${\mathbf m}_t$ are additionally fed into the gate by following the update computations in Eqs. \eqref{eq:reset_gate}-\eqref{eq:hidden}, where $\varphi$ is a non-linear activation function. {\chg It is worth noting that we learn the corresponding parameter for $\boldsymbol{\alpha}_t$, thus allowing the model to learn the extent to which the attention weights are reflected in the input.} In addition, the missing patterns are also modeled by directly feeding a masking vector and are linearly combined with other internal representations.

\subsection{Mortality Prediction}
\label{subsec:Mortality Prediction}
To predict the probability of \emph{in-hospital mortality}, we use the last GRU hidden state, the representation of which is more powerful than the explicit use of the imputed data because it includes temporal information encoding across all time steps. Given the last hidden state ${\mathbf h}_T$, we apply a fully connected layer, followed by a sigmoid activation function as follows:
\begin{gather}
    p(y=1|{\mathbf h}_T)=\text{sigm}({\mathbf W}_{o}{\mathbf h}_T)
\end{gather}
where 
$\text{sigm}$ is a logistic sigmoid activation function and ${\mathbf W}_{o}$ is a parameter from a classifier. For simplicity, the recurrence, uncertainty-gated attention, and prediction model parameters are summarized as $\psi=\{{\mathbf W}_{\{{\gamma}, \beta, x, \alpha, r, u, h, o\}}, {\mathbf U}_{\{r,u,h\}}, {\mathbf V}_{\{r,u,h\}}, {\mathbf b}_{\{\gamma, \beta, x, \alpha, r,u,h\}}\}$.
The overall algorithm for a unidirectional VRNN is described in Algorithm \ref{alg:uniVRNN}.

%%%%%%%%%%%%%%%%%%%%%%%%%%%%%%%%%%%%%%%%%%%%%%
\subsection{Bidirectional VRNN}
\label{subsec:Bidirectional VRNN}
To better capture the long-term dependency of clinical records, in this study, we exploit \emph{bidirectional recurrent dynamics}. In other words, given a time series $\{\hat{\mathbf{x}}_1, \hat{\mathbf{x}}_2, ..., \hat{\mathbf{x}}_T\}$ from the forward direction, each time series can also be derived from the backward direction function, \ie, $\{\hat{\mathbf{x}}'_1, \hat{\mathbf{x}}'_2, ..., \hat{\mathbf{x}}'_T\}$. The final estimated imputation is the mean of $\hat{\mathbf{x}}_T$ and $\hat{\mathbf{x}}'_T$ at time $t$, and the prediction is based on the average logit from a forward and backward VRNN. The final loss is obtained by accumulating the forward loss $\{l_1, l_2, ..., l_T\}$ and the backward loss $\{l'_1, l'_2, ..., l'_T\}$.
This bidirectional approach helps solve the problem of inefficient training and biased explosions caused by error delays that often occur in the time sequence.

%%%%%%%%%%%%%%%%%%%%%%%%%%%%%%%%%%%%%%%%%%%%%%
\subsection{Learning}
\label{subsec:Learning}
To train all model parameters \{$\phi$, $\theta$, $\psi$\}, we use a composite objective function, which consists of (i) VRNN loss $\mathcal{L}_{\text{VRNN}}$, (ii) consistency loss $\mathcal{L}_{\text{cons}}$, (iii) masked value imputation loss $\mathcal{L}_{\text{imp}}$, and (iv) classification loss $\mathcal{L}_{\text{cls}}$. The VRNN loss is calculated by accumulating the reconstruction error and KL divergence over the time-series for each sample as follows:
\begin{equation}
\begin{array}{l}
\displaystyle
	l_t^{(n)}=\mathbb{E}_{q(\bz_{1:T}|{\tilde\bx}_{1:T}^{(n)};\phi)}\big[\log p({\tilde\bx}_t^{(n)}|\bz_{1:t},\tilde\bx_{1:t-1}^{(n)})-\\
	%\text{KL}(q(\bz_t|\hat{\bx}_{1:T}^{(n)},\bz_{1:t-1};\phi))||p(\bz_t|\bx_{1:t}^{(n)},\bz_{1:t-1};\theta)\big]\\
	\textit{KL}\infdivx{q(\bz_t|{\tilde\bx}_{1:T}^{(n)},\bz_{1:t-1};\phi)}{p(\bz_t|\tilde\bx_{1:t}^{(n)},\bz_{1:t-1};\theta)}\big].\label{eq:unsup}
	%\newcommand{\infdiv}{D_{KL}\infdivx}
	%\infdiv{q(\bz_t|\hat{\bx}_{1:T}^{(n)},\bz_{1:t-1};\phi)}{p(\bz_t|\bx_{1:t}^{(n)},\bz_{1:t-1};\theta)}\big].
\end{array}
\end{equation}
All losses for $N$ samples are gathered, which leads to $\mathcal{L}_{\text{VRNN}}=\sum_{n=1}^N\sum_{t=1}^T l_t^{(n)}$. The consistency loss $\mathcal{L}_{\text{cons}}$ is defined as the difference between the forward estimation $\hat{\mathbf{x}}_t$ and backward estimation $\hat{\mathbf{x}}'_t$ over time through the mean absolute error (MAE) for the consistent estimations of both the forward and backward directions. In terms of the imputation loss $\mathcal{L}_{\text{imp}}$, we 
calculate \emph{masked MAE} between the original sample $\bX$ as the ground truth and the imputed sample $\hat{\mathbf{X}}$ for the initially marked values only\footnote{Actually, we randomly selected 10\% of non-missing values and removed them to investigate the masked imputation loss. Thus, another masking vector $\mathbf{M}_{\text{imp}}$ is introduced for the purpose of marking the selected values.} by $\bM_{\text{imp}}$ as:
\begin{gather}
	\mathcal{L}_{\text{imp}} = \frac{1}{N} \sum_{n=1}^{N}\left|{\bX}^{(n)}\odot{\bM}_{\text{imp}}^{(n)}-\hat{\mathbf{X}}^{(n)}\odot{\bM}_{\text{imp}}^{(n)}\right|
\end{gather}

Moreover, in this study, we address the poor classification problem from highly imbalanced data frequently found in a healthcare dataset. To accurately detect minority class observations that are often crucial (\ie, in-hospital death), we exploit a focal loss \citep{lin2017focal}. The loss formulation is similar to the standard cross entropy loss, but reshaped such that it down-weights the loss assigned to well-classified examples in the following way:

\begin{align}
	\label{eq:focal_loss}
	\mathcal{L}_{\text{cls}} &= \sum_{n=1}^{N}- \omega_1 (1-\hat{y}^{(n)})^{\omega_2} \log (\hat{y}^{(n)})\\
	\hat{y}^{(n)} &= 
		\begin{cases} 
			p^{(n)} 	& \text{if } y^{(n)} = 1 \\
			1-p^{(n)} 	& \text{otherwise}
		\end{cases}
\end{align}
where $p^{(n)}$ is the model's predicted probability for $y^{(n)}= 1$, $\omega_1\in[0,1]$ is a weighting factor used to balance the importance between them, and $\omega_2\geq0$ is a focusing parameter applied to focus on the minority class.

Hence, all losses are then accumulated by integrating both the forward loss and backward loss, defining the composite loss as $\mathcal{L}=\lambda_{\text{VRNN}}(\mathcal{L}_{\text{VRNN}}+\mathcal{L}'_{\text{VRNN}})+\lambda_{\text{cons}}(\mathcal{L}_{\text{cons}}+\mathcal{L}'_{\text{cons}})+\lambda_{\text{imp}}(\mathcal{L}_{\text{imp}}+\mathcal{L}'_{\text{imp}})+(\mathcal{L}_{\text{cls}}+\mathcal{L}'_{\text{cls}})$, where $\lambda_{\text{VRNN}}$, $\lambda_{\text{cons}}$ and $\lambda_{\text{imp}}$ are hyper-parameters that control the ratio between the losses. We optimize all the parameters of our model in an end-to-end manner via this composite loss. 

\begin{algorithm}[h]
\caption{{Uncertainty-gated stochastic recurrent imputation in unidirectional VRNN}}
\label{alg:uniVRNN}
\begin{algorithmic}[1]
\Require {$\tilde{\bx}_{1:T}$, $\mathbf{m}_{1:T}$, $\Delta\theta_0\leftarrow 0$, $\Delta\phi\leftarrow 0$, $\Delta\theta\leftarrow 0$, $\Delta\psi\leftarrow 0$,
$\mathcal{L}\leftarrow0$}
	\Repeat
  	 \For{$t\leftarrow1$ to $T$}
     	 \State $p(\bz_t;{{\theta_{0}}})\leftarrow\mathcal{N}\left(\boldsymbol{\mu}_{0,t}, \text{diag}(\boldsymbol{\sigma}_{0,t}^2)\right)$
	 	\State $q({\mathbf z}_t|\tilde{\mathbf x}_t;{{\phi}})\leftarrow\mathcal{N}\left(\boldsymbol{\mu}_{\bz,t}, \text{diag}(\boldsymbol{\sigma}_{\bz,t}^2)\right)$ %\eqref{eq:inference_1}-\eqref{eq:inference_2}
		\State $\boldsymbol{\epsilon}\sim\mathcal{N}(\mathbf{0},\mathbf{I})$
		 \State $\bz_t\leftarrow\boldsymbol{\mu}_{\bz,t}+\boldsymbol{\sigma}_{\bz,t}\odot\boldsymbol{\epsilon}$
	 	\State $p(\tilde{\mathbf x}_t|{\mathbf z}_t;{{\theta}})\leftarrow\mathcal{N}\left(\boldsymbol{\mu}_{\bx,t}, \text{diag}({\boldsymbol{\sigma}^2_{\bx,t}})\right)$
	 	\State $\boldsymbol{\gamma}_t\leftarrow\exp\{-\max(\mathbf{0}, \mathbf{W}_{\gamma}\boldsymbol{\delta}_t+\mathbf{b}_{\gamma})\}$
	 	\State $\boldsymbol{\beta}_t\leftarrow\sigma\left(\mathbf{W}_{\beta}[\boldsymbol{\gamma}_t \circ \mathbf{m}_t]+\mathbf{b}_{\beta}\right)$
	 	\State $\mathbf{c}_t\leftarrow\boldsymbol{\beta}_t \odot\tilde{\bx}_t+(\mathbf{1}-\boldsymbol{\beta}_t)\odot  \boldsymbol{\mu}_{\bx,t}$
	 	\State $\tilde{\mathbf{c}}_t\leftarrow\mathbf{W}_{x}\boldsymbol{\mu}_{\bx,t}+\mathbf{b}_{x}$
		\State $\hat{\mathbf{c}}_t\leftarrow\text{max-pool}\left(\mathbf{c}_t*\tilde{\mathbf{c}}_t\right)$
		\State $\hat{\bx}_t\leftarrow \mathbf{m}_t \odot\tilde{\bx}_t+(\mathbf{1}-\mathbf{m}_t)\odot  \hat{\mathbf{c}}_t$
		\State ${\mathbf{u}}_t\leftarrow({\mathbf{1}}-\mathbf{m}_t)\odot{\boldsymbol{\sigma}}_{\bx,t}$
		\State $\boldsymbol{\alpha}_t\leftarrow \exp\{-\max(\mathbf{0},{\mathbf W}_{\alpha}{\mathbf{u}_t}+{\mathbf b}_{\alpha})$
		\State $\mathbf{h}_t \leftarrow \text{GRU}\left(\mathcal{F}^{\bx}(\hat{\mathbf{x}}_t),\mathcal{F}^{\bz}(\bz_t),\mathbf{h}_{t-1}, \mathbf{u}_t\right)$
% 		 \State $l_t\leftarrow \text{Eq. (24)}$
     	\EndFor
	\State $\mathcal{L}\leftarrow \lambda_{\text{VRNN}}\mathcal{L}_{\text{VRNN}}+\lambda_{\text{cons}}\mathcal{L}_{\text{cons}}+\lambda_{\text{imp}}\mathcal{L}_{\text{imp}}+\mathcal{L}_{\text{cls}}$
	\State $\Delta\theta_0\leftarrow \frac{\partial}{\partial\theta_0}\mathcal{L}$
	\State $\Delta\phi\leftarrow \frac{\partial}{\partial\phi}\mathcal{L}$
	\State $\Delta\theta\leftarrow \frac{\partial}{\partial\theta}\mathcal{L}$
	\State $\Delta\psi\leftarrow \frac{\partial}{\partial\psi}\mathcal{L}$
	\State $\theta_0\leftarrow \text{Optimize}(\theta_0,\Delta\theta_0)$
	\State $\phi\leftarrow \text{Optimize}(\phi,\Delta\phi)$
	\State $\theta\leftarrow \text{Optimize}(\theta,\Delta\theta)$
	\State $\psi\leftarrow \text{Optimize}(\psi,\Delta\psi)$
	\Until{Convergence}
  \end{algorithmic}
\end{algorithm}

\begin{table*}[tb]
\setlength{\tabcolsep}{12pt}
\renewcommand*{\arraystretch}{1.05}
\caption{Performances of the mortality prediction task ($\textit{mean $\pm$ std}$ from 5-cross validation)}
\begin{center}
\begin{tabular}{cp{1.65cm}cccccc}
\toprule
{\bf Masking ratio}     & {\bf Method} 	& \multicolumn{2}{c}{\bf MIMIC-III}	& \multicolumn{2}{c}{\bf PhysioNet} 	\\
                        &               &  {\bf AUC}      & {\bf AUPRC}       &  {\bf AUC}      & {\bf AUPRC}\\\midrule
\multirow{9}{*}{\bf 5\%}& {\bf GRU-Zero}& 0.807 $\pm$ 0.021 & 0.345 $\pm$ 0.036 & 0.819 $\pm$ 0.021 & 0.444 $\pm$ 0.033 \\
	                    & {\bf GRU-Mean}& 0.796 $\pm$ 0.018 & 0.330 $\pm$ 0.016 & 0.793 $\pm$ 0.021 & 0.431 $\pm$ 0.041	\\
	                    & {\bf GRU-KNN}	& 0.794 $\pm$ 0.020 & 0.338 $\pm$ 0.017 & 0.790 $\pm$ 0.026 & 0.412 $\pm$ 0.049		\\
                        & {\bf GRU-D \citep{che2018recurrent}}   & 0.853 $\pm$ 0.013          & 0.380 $\pm$ 0.019   & 0.809 $\pm$ 0.030          & 0.464 $\pm$ 0.046		\\
                        & {\bf M-RNN \citep{yoon2017multi}} & 0.822 $\pm$ 0.010          & 0.317 $\pm$ 0.016    & 0.781 $\pm$ 0.023 & 0.383 $\pm$ 0.042		\\
                        & {\bf RITS \citep{cao2018brits}} & 0.854 $\pm$ 0.009          & 0.373 $\pm$ 0.028  & 0.810 $\pm$ 0.015     & 0.456 $\pm$ 0.039		\\
                        & {\bf BRITS \citep{cao2018brits}} & 0.863 $\pm$ 0.014      & 0.414 $\pm$ 0.025 & 0.824 $\pm$ 0.004 & 0.460 $\pm$ 0.042 		\\
                        & {\bf SAnD \citep{song2018attend}} & 0.830 $\pm$ 0.010	 	& 0.374 $\pm$ 0.024	& 0.787 $\pm$ 0.017 & 0.426 $\pm$ 0.026	             \\
	                    & {\bf Ours} & {\bf0.865 $\pm$ 0.008}  & {\bf 0.416 $\pm$ 0.029} & {\bf 0.832 $\pm$ 0.018} & {\bf 0.470 $\pm$ 0.054}		\\\hline
\multirow{9}{*}{\bf 10\%}& {\bf GRU-Zero}	    & 0.809 $\pm$ 0.023 & 0.346 $\pm$ 0.024 & 0.822 $\pm$ 0.022 & 0.440 $\pm$ 0.039	\\
	                    & {\bf GRU-Mean}	    & 0.795 $\pm$ 0.011 & 0.326 $\pm$ 0.020 & 0.793 $\pm$ 0.018 & 0.436 $\pm$ 0.028	\\
	                    & {\bf GRU-KNN}		    & 0.798 $\pm$ 0.015 & 0.328 $\pm$ 0.021 & 0.788 $\pm$ 0.021 & 0.428 $\pm$ 0.041	\\
                        & {\bf GRU-D \citep{che2018recurrent}}  & 0.853 $\pm$ 0.013     & 0.380 $\pm$ 0.019 & 0.809 $\pm$ 0.030          & 0.464 $\pm$ 0.046 		\\
                        & {\bf M-RNN \citep{yoon2017multi}} & 0.824 $\pm$ 0.006          & 0.322 $\pm$ 0.023    & 0.781 $\pm$ 0.023 & 0.383 $\pm$ 0.042		\\
                        & {\bf RITS \citep{cao2018brits}} & 0.860 $\pm$ 0.004   & 0.392 $\pm$ 0.004 & 0.810 $\pm$ 0.015 & 0.456 $\pm$ 0.039		\\
                        & {\bf BRITS \citep{cao2018brits}} & 0.864 $\pm$ 0.011           & 0.412 $\pm$ 0.032 & 0.824 $\pm$ 0.004 & 0.460 $\pm$ 0.042		\\
                        & {\bf SAnD \citep{song2018attend}} & 0.826 $\pm$ 0.009	 & 0.372 $\pm$ 0.025		& 0.787 $\pm$ 0.017     & 0.426 $\pm$ 0.026	 \\
            	        & {\bf Ours} & {\bf 0.865 $\pm$ 0.010} & {\bf 0.415 $\pm$ 0.014}   & {\bf 0.829 $\pm$ 0.022}  & {\bf 0.465 $\pm$ 0.054}		\\\toprule
\end{tabular}
\label{tb:prediction_result}
\end{center}
\end{table*}

%%%%%%%%%%%%%%%%%%%%%%%%%%%%%%%%%%%%%%%%%%%%%%%%%%%%%%%%%%%%%%%%%%%%%%%%%%%%%%%%%%%%%%%%%%%%
\section{Experiments} 
In this section, we evaluate the proposed uncertainty-gated stochastic sequential imputation method on in-hospital mortality prediction and missing value imputation task on two publicly available healthcare dataset, {\chg (i) Medical Information Mart for Intensive Care III (MIMIC-III) and (ii) PhysioNet Challenge 2012}, which have multivariate time-series that include numerous missing values. {\chg To compare the results depending on the ratio of missing values, we considered two masking scenarios where 5\% or 10\% of the observations were additionally masked for each dataset.}

We reported the performances of the mortality prediction task with the average results from a 5-fold cross validation in terms of (i) the area under the ROC curve (AUC) and (ii) the area under the precision-recall curve (AUPRC). The results of the missing value imputation are reported in terms of the MAE. Here, with respect to the mortality prediction task, both the prediction and imputation are conducted during training, although only prediction is applied during testing. In addition, regarding the imputation task, only a missing value imputation is conducted during the training and testing. 

We compared the results of the two tasks with other state-of-the-art methods in the literature to show the superiority of the proposed method. In addition, we conducted extensive ablation studies of our model to evaluate the effects of different components in the proposed method.

In addition to validating the performances of the two tasks, we visualized the imputation estimates with the uncertainties predicted from our model against actual observations over time, and further investigated the behavior of our model regarding the representation of the uncertainties for the imputed estimates in terms of the correlation between imputation MAE and uncertainty.

All the codes are available at ``https://open-after-acceptance”.

\subsection{Data}
\label{subsec:Data}
{\chg We used two publicly available datasets, namely, the MIMIC-III and PhysioNet challenge 2012 datasets.}

%%%%%%%%%%%%%%%%%%%%%%%%%%%%%%%%%%%%%%%%%%%%%%
\subsubsection{MIMIC-III}
\label{subsubsec:MIMIC-III}
We used the publicly available real-world EHR dataset, MIMIC-III\footnote{Available at \url{https://mimic.physionet.org/}.}, which contains longitudinal measurements for more than 40,000 critical care patients. We selected  a subset of 13,998 patients with at least 48 hours of hospital stay, and sampled the time-series every 2 hours in the first 48 hours. For each patient, 99 different longitudinal measurements were selected, which were divided into four main categories: laboratory measurements, inputs to patients, outputs collected from patients, and drug prescriptions. The selected time series were scarcely observed leading to a missing rate of approximately 93.92\%. For the in-hospital mortality label, the ratio between 1,181 positive (dead in hospital) and 12,817 negative (alive in hospital) was approximately 1:10.8. 

% %%%%%%%%%%%%%%%%%%%%%%%%%%%%%%%%%%%%%%%%%%%%%%
% \subsubsection{eICU}
% \label{subsubsec:eICU}
% {\todo}

%%%%%%%%%%%%%%%%%%%%%%%%%%%%%%%%%%%%%%%%%%%%%%
\subsubsection{PhysioNet challenge 2012}
\label{subsubsec:PhysioNet challenge 2012}
We also used the PhysioNet challenge 2012 dataset\footnote{Available at \url{https://physionet.org/content/challenge-2012/1.0.0/}.}, which contains longitudinal measurements for 4,000 critical care patients with at least 48 hours of hospital stay. Here, we removed 3 patients from original dataset who had no observations at all. We sampled the observations hourly in the first 48 hours, taking the mean value of multiple observations within one hour. For each patient, 35 different longitudinal measurements were exploited, 
including the time-series measurements of vital signs and lab test results. The time-series data contain a large number of missing values with a missing rate of approximately 80.51\% and an in-hospital mortality label imbalanced at a ratio of approximately 1:6 between 554 in-hospital deaths and 3,443 survival cases.

%%%%%%%%%%%%%%%%%%%%%%%%%%%%%%%%%%%%%%%%%%%%%%
\subsection{Preprocessing and Training}
\label{subsec:Preprocessing_and_Training}
For the MIMIC-III dataset, data cleaning was conducted by handling inconsistent units, multiple recordings made at the same time, and the range of the recorded feature values. We referred to \cite{che2018recurrent, purushotham2018benchmarking} for feature selection, data cleaning, and preprocessing.

For all datasets, because each variable has a different range, all inputs were first Winsorized for removing outliers and then $z$-normalized using the global mean and standard deviation from the entire training set to achieve a zero mean and unit variance in a variable-wise manner, as described in \cite{bahadori2019temporal}.

We trained our models using the Rectified Adam (RAdam) optimizer \citep{liu2019variance} with an initial learning rate of $0.001$ and a multiplicative decay of 0.5 for $80$ epochs using mini-batches of {$64$} samples. We chose the final optimal model based on the performance of the validation set.

%%%%%%%%%%%%%%%%%%%%%%%%%%%%%%%%%%%%%%%%%%%%%%
\subsection{Model Implementations}
\label{subsec:Model_Implementations}
The VRNN comprises an inference network, a generative network, feature extractors, and an RNN, which are built using neural networks in our implementation. The inference and generative network were fully connected with 2 hidden layers that are a linear operation followed by batch normalization, and a rectified linear unit (ReLU) activation, where the dimensions of the latent variables are 32 and 16 for each layer, respectively. For the feature extractors, we constructed a single hidden layer using a Tanh activation to extract complex non-linear features. For the RNN, a single layer with 64 GRU hidden units was employed with a Tanh activation function. The $\omega_1$ and $\omega_2$ in the focal loss were chosen to be 5 and 0.25, respectively. The ratio in the composite loss was set to $\lambda_{\text{VRNN}}=1e^{-5}$, $\lambda_{\text{cons}}=1$ and $\lambda_{\text{imp}}=1e^{-2}$ as a result of varying their values in [0, $5e^{-6}$, $1e^{-5}$, $1e^{-4}$, $1e^{-3}$, $1e^{-2}$, $1e^{-1}$, 1].

All model parameters were initialized as small random numbers such that their values fell within the standard deviation interval, which is the inverse of the number of input nodes. 

% The ratio in the composite loss: We investigate the effect of the ratio between the losses. Fig. \ref{fig:lamda} presents the classification performance  when $\lambda_2$ is fixed with $1e^{-2}$. Note that we don't show the results of varying the $\lambda_2$, as changing the size of the imputation loss affect very small. Since the values of the focal loss are much smaller than that of unsupervised loss from VRNN, downsizing the unsupervised term contributes to performance improvement, especially with the $\lambda$ value of $1e^{-5}$.

%%%%%%%%%%%%%%%%%%%%%%%%%%%%%%%%%%%%%%%%%%%%%%
\subsection{Baseline Methods}
\label{subsec:Baseline_Methods}
We validated the efficacy of our framework by dividing the evaluation of (i) the in-house mortality prediction task  and (ii) the missing value imputation task. Regarding the mortality prediction task, we compared our proposed method with the vanilla GRU with the zero, mean, and KNN imputation\footnote{
The Mean Impute (SimpleFill) and KNN are implemented by using fancyimpute library in Python. The code is publicly available at \url{https://github.com/iskandr/fancyimpute}.} (\ie, GRU-Zero, GRU-Mean, and GRU-KNN); RNN-based SOTA models such as GRU-D \citep{che2018recurrent}, M-RNN \citep{yoon2017multi}, BRITS and RITS \citep{cao2018brits}, removing the backward direction in BRITS; and a transformer-based SOTA model, SAnD \citep{song2018attend}, which employs a masked self-attention mechanism for clinical diagnosis.

% In the meantime, as the self-attention mechanism \citep{vaswani2017attention} without any recurrence surged in the sequence-to-sequence modeling, \citep{song2018attend} \etal\ employed a masked self-attention mechanism via positional encoding and dense interpolation strategies for clinical time series modeling, and showed reasonable performance in the EHR multi-tasks.

% We validate the efficacy of our framework by dividing the evaluation of (i) in-house mortality prediction task  and (ii) missing value imputation task. Regarding the mortality prediction task, we compared our proposed method with GRU-D \citep{che2018recurrent}, M-RNN \citep{yoon2017multi}, BRITS and RITS, removing backward direction in BRITS \citep{cao2018brits}, GRU-I \citep{luo2018multivariate}, Multi-task LSTM \citep{harutyunyan2019multitask} and SAnD (Simply Attend and Diagnose) \citep{song2018attend} that employs the self-attention widely used in natural language processing. 

For a missing value imputation, we included Zero Impute, Mean Impute, KNN, GRU-D \citep{che2018recurrent}, M-RNN \citep{yoon2017multi}, BRITS, and RITS \citep{cao2018brits}. The prediction tasks for all datasets are compared in Table \ref{tb:prediction_result}, and that of the imputation task is compared in Table \ref{tb:imputation_result}.

%%%%%%%%%%%%%%%%%%%%%%%%%%%%%%%%%%%%%%%%%%%%%%
\section{Results}
\label{sec:Results}

%%%%%%%%%%%%%%%%%%%%%%%%%%%%%%%%%%%%%%%%%%%%%%
\subsection{Result of Mortality Prediction}
\label{subsec:Mortality_Prediction_Results}
Table \ref{tb:prediction_result} compares the results of our proposed method with those of the baselines for mortality prediction. In both masking scenarios, our model achieved the best classification performance on both datasets. The results of a relatively simple imputation method with a GRU (GRU-Zero, GRU-Mean, and GRU-KNN) suggest the need for a more sophisticated imputation method compared with the other SOTA models. Among the RNN-based baselines, BRITS demonstrated a competitive performance. In contrast, SAnD showed a relatively low performance compared with the other RNN-based methods, despite the benefits of computational efficiency. These experimental results validate the efficacy of our proposed method equipped with the stochastic recurrent imputation using the VRNN and GRU-U cell, showing its superior performance in the downstream task.

\begin{table}[tb]
\setlength{\tabcolsep}{7pt}
\renewcommand*{\arraystretch}{1.05}
\caption{Results of missing value imputations measured by MAE score}
\begin{center}
\begin{tabular}{cp{1.65cm}cccc}
\toprule
{\bf Masking ratio}     & {\bf Method} 	& {\bf MIMIC-III}	        & {\bf PhysioNet} 	   \\\midrule
\multirow{9}{*}{5\%}    & Zero Impute	& 0.724 $\pm$ 0.005         & 0.788 $\pm$ 0.006		\\
	                    & Mean Impute	& 0.520 $\pm$ 0.003		    & 0.510 $\pm$ 0.010		\\
	                    & KNN			& 0.508 $\pm$ 0.003		    & 0.396 $\pm$ 0.005		\\
                        & GRU-D \citep{che2018recurrent}    	& 0.584 $\pm$ 0.007 	    & 0.660 $\pm$ 0.014		\\
                        & M-RNN \citep{yoon2017multi}     	& 0.451 $\pm$ 0.008 	    & 0.411 $\pm$ 0.011		\\
                        & RITS \citep{cao2018brits}    	    & 0.354 $\pm$ 0.005 	    & 0.325 $\pm$ 0.007		\\
                        & BRITS	\citep{cao2018brits}	    & {\bf 0.332 $\pm$ 0.005} 	& {\bf 0.297 $\pm$ 0.008} \\
                        % GRU-I 		& 	 		                &			             \\\cline{1-3}
	                    & Ours		    & 0.497 $\pm$ 0.012	 	    & 0.525 $\pm$ 0.004 \\\hline
\multirow{9}{*}{10\%}   & Zero Impute	& 0.724 $\pm$ 0.005		    & 0.791 $\pm$ 0.004		\\
	                    & Mean Impute	& 0.523 $\pm$ 0.006	        & 0.513 $\pm$ 0.004		\\
	                    & KNN			& 0.515 $\pm$ 0.004	        & 0.415 $\pm$ 0.004		\\
                        & GRU-D \citep{che2018recurrent}    	& 0.582 $\pm$ 0.006		    & 0.658 $\pm$ 0.016		\\
                        & M-RNN \citep{yoon2017multi}     	& 0.438 $\pm$ 0.005	 	    & 0.397 $\pm$ 0.009		\\
                        & RITS \citep{cao2018brits}    	    & 0.331 $\pm$ 0.006		    & 0.308 $\pm$ 0.006		\\
                        & BRITS	\citep{cao2018brits}	    & {\bf 0.312 $\pm$ 0.004}	& {\bf 0.283 $\pm$ 0.008} \\
                        % GRU-I 		& 	 		                &			             \\\cline{1-3}
            	        & Ours		    & 0.503 $\pm$ 0.011	 		& 0.526 $\pm$ 0.010  \\\toprule
\end{tabular}
\label{tb:imputation_result}
\end{center}
\end{table}

%%%%%%%%%%%%%%%%%%%%%%%%%%%%%%%%%%%%%%%%%%%%%%
\subsection{Result of Missing Value Imputation}
\label{subsec:Result_of_Missing_Value_Imputation}
Table \ref{tb:imputation_result} compares other imputation baselines for the missing value imputation task.
Whereas BRITS showed the lowest MAE scores under two masking scenarios on both datasets, the MAE scores of our proposed method are slightly higher, comparable to those of Mean Impute, and in most cases, better than those of Zero Impute, Mean Impute, KNN, and GRU-D.

\begin{figure*}[th!]
% 	\centering
% 	\begin{subfigure}[b]{\linewidth}\centering
% % 		\includegraphics[width=\columnwidth]{./figures/var_2_batch_0_fold_0_epoch_1.png}
% 		\caption{Imputation results on MIMIC-III dataset in a 5\% scenario.}
% 		\label{fig:Vis_Imp_MIMIC_5}
% 	\end{subfigure}
% 	\centering
% 	\begin{subfigure}[b]{\linewidth}\centering
% % 		\includegraphics[width=\columnwidth]{./figures/var_2_batch_0_fold_0_epoch_1.png}
% 		\caption{Imputation results on MIMIC-III dataset in a 10\% scenario.}
% 		\label{fig:Vis_Imp_MIMIC_10}
% 	\end{subfigure}
	\centering
	\begin{subfigure}[b]{0.88\linewidth}\centering
		\includegraphics[width=\columnwidth]{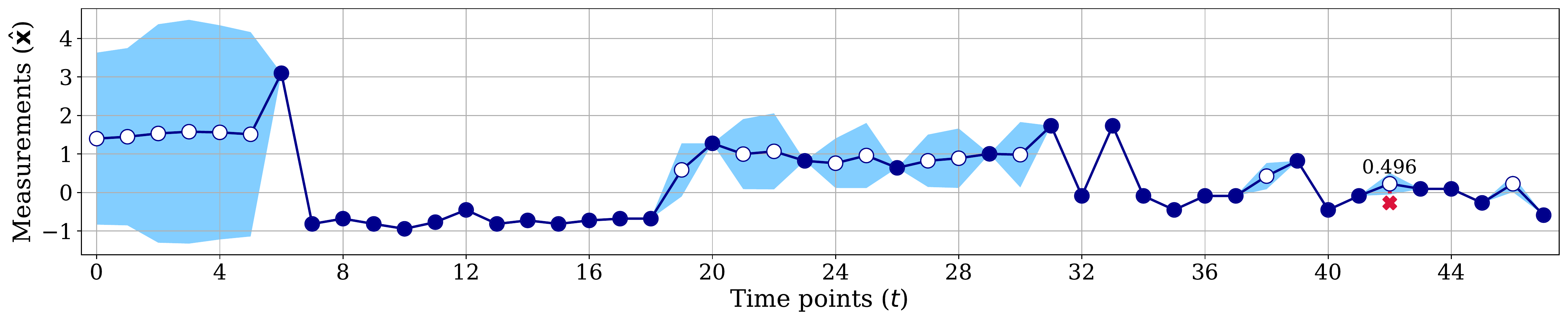}
		\caption{}
% 		\caption{Imputation result on PhysioNet dataset in a 5\% masking scenario.}
		\label{fig:Vis_Imp_PhysioNet_5}
	\end{subfigure}
	\centering
	\begin{subfigure}[b]{0.88\linewidth}\centering
		\includegraphics[width=\columnwidth]{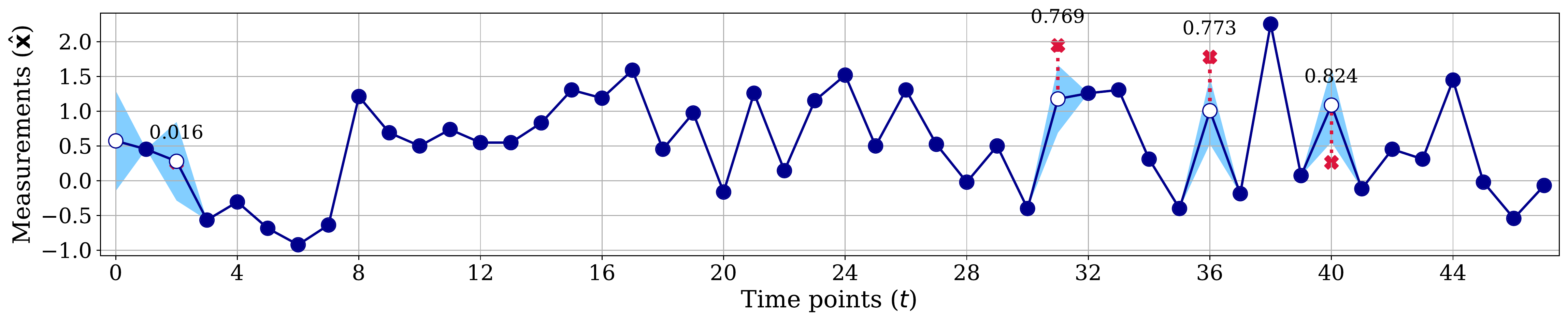}
		\caption{}
% 		\caption{Imputation result on PhysioNet dataset in a 10\% masking scenario.}
		\label{fig:Vis_Imp_PhysioNet_10}
	\end{subfigure}
	\caption{Visualization of imputation estimates for PhysioNet dataset under both (a) 5\% and (b) 10\% masking scenarios. Blue filled dots are observed measurements, blue lines blue shades with hollow dots are the imputations and uncertainties estimated from our model, respectively, and red x marks are masked ground-truth observations. In addition, the number above the red dashed vertical line represents the exact MAE values between masked ground-truth observation values and model predictions.}
	\label{fig:imputation_overall}
%	\label{fig:overall}
\end{figure*}

\begin{table}[t!]
\caption{Performance of a set of ablation experiments for mortality prediction task}
\begin{center}
\begin{tabular}{p{3.3cm}ccc}
\toprule
    	{\bf Method} 		& {\bf AUC} 			    & {\bf AUPRC}  \\\midrule
     	VAE + vanilla GRU   & 0.759 $\pm$ 0.007		    & 0.373 $\pm$ 0.019 \\
     	VAE + GRU-U      	& 0.813 $\pm$ 0.017		    & 0.442 $\pm$ 0.029    \\
     	VRNN + vanilla GRU  & 0.794 $\pm$ 0.013		    & 0.411 $\pm$ 0.038	 \\
     	VRNN + GRU-U		& {\bf 0.832 $\pm$ 0.018} 	& {\bf 0.470 $\pm$ 0.054}  \\\toprule
\end{tabular}
\label{tb:ablation}
\end{center}
\end{table}

\begin{table}[t!]
\caption{Performance of the ablation experiments related to loss and ${\mathbf W}_{\alpha}$ for the mortality prediction task}
\begin{center}
\scalebox{1.1}{
\begin{tabular}{p{0.9cm}p{1.4cm}cc}
\toprule
{\bf Ablation}          & {\bf Method} 		& {\bf AUC} 		        & {\bf AUPRC}  \\\midrule
\multirow{2}{*}{Loss}   & BCE loss			& 0.776 $\pm$ 0.014 	    & 0.384 $\pm$ 0.032  \\
                        & Focal loss      	& {\bf 0.832 $\pm$ 0.018} 	& {\bf 0.470 $\pm$ 0.054} \\ \hline
\multirow{2}{*}{${\mathbf W}_{\alpha}$}   & Full    & 0.829 $\pm$ 0.016		    & {\bf 0.472 $\pm$ 0.035}  \\
                        & Diagonal    		& {\bf 0.832 $\pm$ 0.018} & 0.470 $\pm$ 0.054 \\\toprule       
  %   	VRNN + vanilla GRU     	& 0.794 $\pm$ 0.013		& 0.411 $\pm$ 0.038	 \\\cline{1-3}
   %  	VRNN + GRU-U		& {\bf 0.823 $\pm$ 0.015}	& {\bf 0.458 $\pm$ 0.042}  \\\cline{1-3}
\end{tabular}
\label{tb:ablation}
}
\end{center}
\end{table}

% \begin{table}[h!]
% \caption{Performance comparison of the ablation experiment related to the prediction layer measured by AUC and AUPRC scores on the mortality prediction task.}
% \begin{center}
% \begin{tabular}{p{3.3cm}ccc}
% \toprule
% 	{\bf Method} 				& {\bf AUC} 	    & {\bf AUPRC}  \\\midrule
%     Attention mechanism      	& 0.811 $\pm$ 0.020     & 0.444 $\pm$ 0.036  \\
%     Mean pooling mechanism	    & 0.814 $\pm$ 0.013     & 0.444 $\pm$ 0.037	\\
%     Last hidden state     		& {\bf 0.832 $\pm$ 0.018} 	& {\bf 0.470 $\pm$ 0.054} \\\toprule
% \end{tabular}
% \label{tb:prediction_layer}
% \end{center}
% \end{table}

% \begin{table}[h!]
% \caption{Performance comparison of the ablation experiment related to  measured by AUC and AUPRC scores on the mortality prediction task.}
% \begin{center}
% \begin{tabular}{p{3.3cm}ccc}
% \toprule
% 	{\bf Method} 			& {\bf AUC} 		        & {\bf AUPRC}  \\\midrule
%     Full weight matrix      & 0.829 $\pm$ 0.016		    & {\bf 0.472 $\pm$ 0.035}  \\
%     Diagonal matrix    		& {\bf 0.832 $\pm$ 0.018} & 0.470 $\pm$ 0.054 \\\toprule
% \end{tabular}
% \label{tb:weight_matrix}
% \end{center}
% \end{table}

\begin{figure*}[th!]
% 	\centering
% 	\begin{subfigure}[b]{0.4\linewidth}\centering
% % 		\includegraphics[width=\columnwidth]{./figures/UNC_MAE.pdf}
% 		\caption{Result on MIMIC-III dataset in a 5\% scenario.}
% 		\label{fig:corr_5_MIMIC}
% 	\end{subfigure}
% 	\hspace{3em}
% 	\centering
% 	\begin{subfigure}[b]{0.4\linewidth}\centering
% % 		\includegraphics[width=\columnwidth]{./figures/UNC_MAE.pdf}
% 		\caption{Result on MIMIC-III dataset in a 10\% scenario.}
% 		\label{fig:corr_10_MIMIC}
% 	\end{subfigure}
	\centering
	\begin{subfigure}[b]{0.39\linewidth}\centering
		\includegraphics[width=\columnwidth]{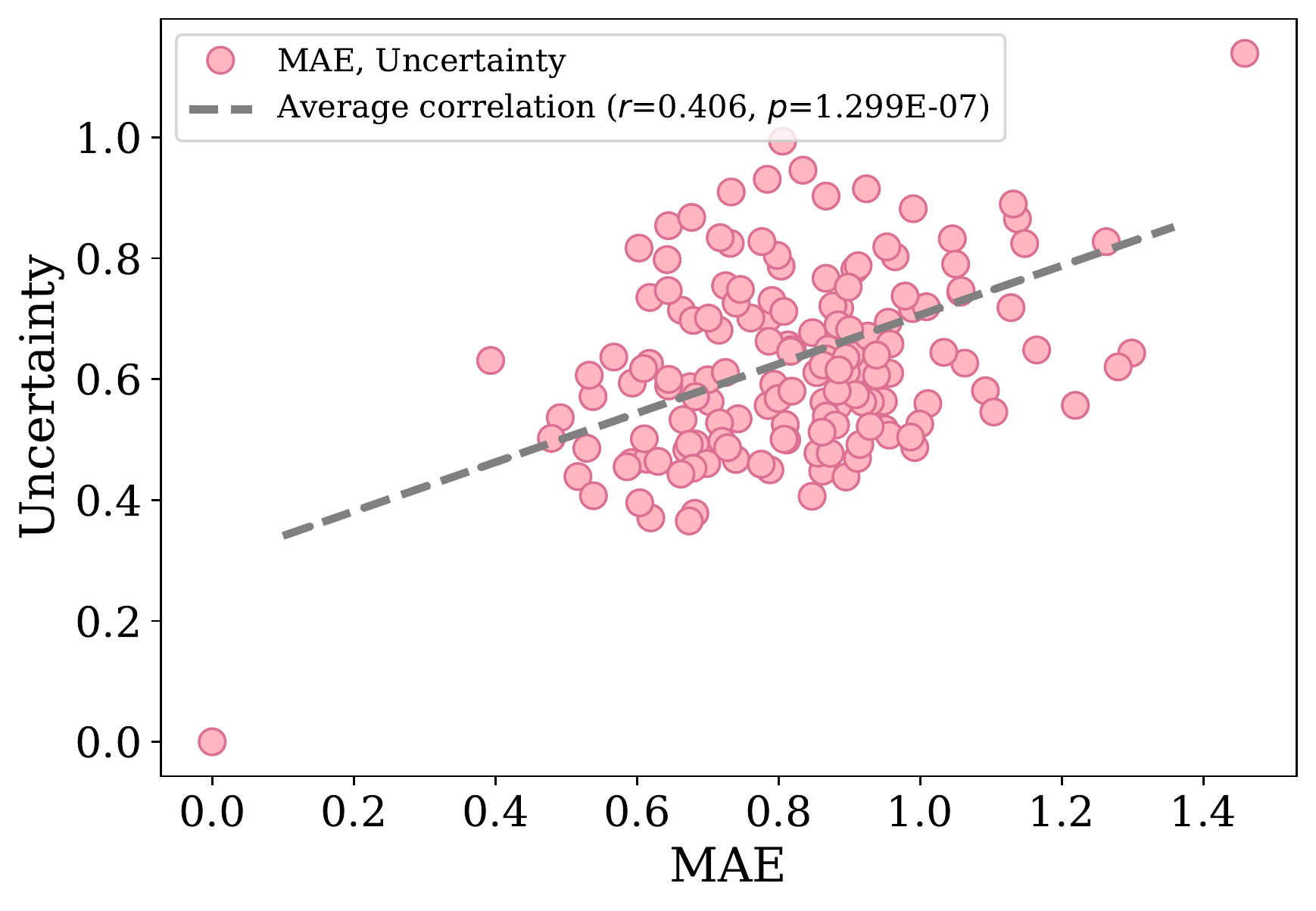}
		\caption{}
% 		\caption{Result on PhysioNet dataset in a 5\% masking scenario.}
		\label{fig:corr_5_PhysioNet}
	\end{subfigure}
	\hspace{4em}
	\centering
	\begin{subfigure}[b]{0.39\linewidth}\centering
		\includegraphics[width=\columnwidth]{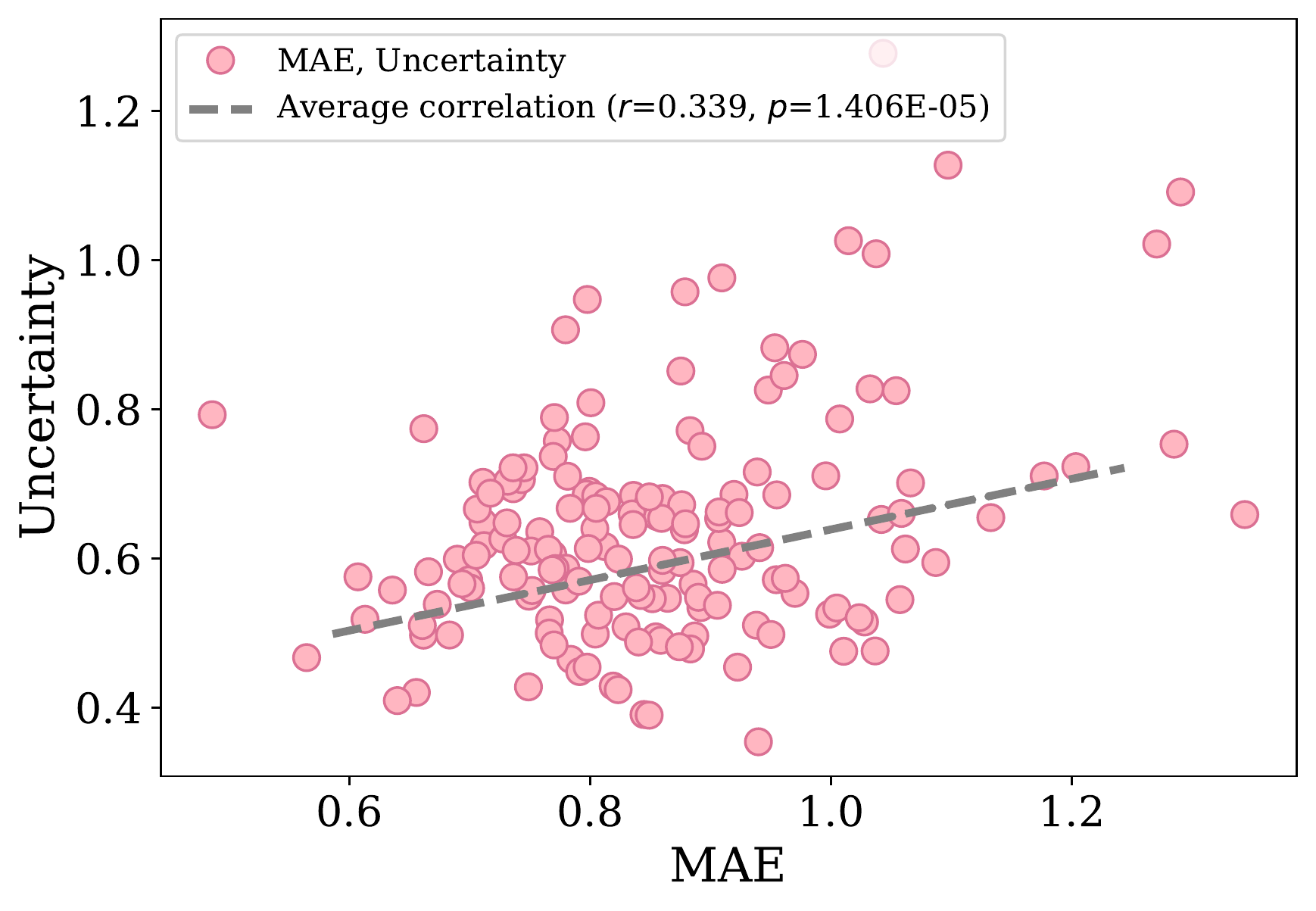}
		\caption{}
% 		\caption{Result on PhysioNet dataset in a 10\% masking scenario.}
		\label{fig:corr_10_PhysioNet}
	\end{subfigure}
	\caption{Visualization of uncertainty values depending on MAE values for PhysioNet dataset under both (a) 5\% and (b) 10\% masking scenarios. Each pink dot represents a pairing of the MAE and the uncertainties for masked observations from a single sample, and the slope of the grey line represents the correlation coefficient.}
	\label{fig:imputation_overall}
%	\label{fig:overall}
\end{figure*}

% %%%%%%%%%%%%%%%%%%%%%%%%%%%%%%%%%%%%%%%%%%%%%%
% \subsection{Comparison of Imputation Error for Different Missing Rates}
% %\label{subsec:Mortality_Prediction_Results}

%%%%%%%%%%%%%%%%%%%%%%%%%%%%%%%%%%%%%%%%%%%%%%
\subsection{Ablation Studies}
\label{subsec:Ablation_Studies}
In this study, we conducted a set of ablation experiments to investigate the influence of different experimental design options of our method, evaluated on the PhysioNet dataset under the 5\% masking scenario. 
\begin{itemize}
	\item Effect of the VRNN: To validate our stochastic recurrent imputation, we compared the case of using a VAE and an RNN separately. Whereas the VAE only considers the dependencies among variables during the imputation process, the VRNN generates a time-series considering the variable and temporal-based dependencies simultaneously. Table \ref{tb:ablation} summarizes the classification performances of two masking scenarios, (VAE+vanilla GRU) \citep{jun2019} versus (VRNN+vanilla GRU) and (VAE+GRU-U) versus (VRNN+GRU-U). Interestingly, the performances improved under both scenarios, particularly when using vanilla GRU with a large margin. Although both approaches provide a stochastic imputation, the experimental results highlight the effectiveness of simultaneously modeling the underlying temporal dependencies and reconstructing the imputation estimates.

	\item Effect of the GRU-U: To verify the effectiveness of our proposed GRU-U cell, we compared the experimental results using the vanilla GRU cell and GRU-U cell. From the prediction results of (VAE+vanilla GRU) versus (VAE+GRU-U) and (VRNN+vanilla GRU) versus (VRNN+GRU-U) in Table \ref{tb:ablation}, we noticed that both results show significant performance improvements; however, in particular, leveraging the GRU-U cell with the VAE is relatively more critical to the predictive task. These experimental findings validate the efficacy of uncertainty-gated attention.
	
	\item Effect of the focal loss: The binary cross-entropy (BCE) loss is widely used in binary classification. By comparing the performance of BCE loss with that of the focal loss in Table \ref{tb:ablation}, we found that the focal loss obtained a better performance than the BCE loss in terms of both AUC and AUPRC. Thus, we can conclude that the focal loss is sufficient for capturing the minority samples in our prediction task.

	\item {\chg Weight matrix ${\mathbf W}_{\alpha}$ in uncertainty-gated attention: The method for propagating uncertainty is determined by ${\mathbf W}_{\alpha}$, where the diagonal matrix effectively makes the decay rate of each variable independent from the others and a full weight matrix makes it dependent. As shown in Table \ref{tb:ablation}, AUPRC was higher when using the full weight matrix, whereas AUC was higher when using the diagonal matrix. This suggests that the dependencies of the medical variables in the calculation of uncertainty-gated attention do not show a significant difference in the performance of the downstream tasks.}
	%Since the values of the focal loss are much smaller than that of unsupervised loss from VRNN, downsizing the unsupervised term helps performance improvement, especially with the $\lambda$ value of $1e^{-5}$.

%$\mathcal{L}=\mathcal{L}_{\text{unsup}}+\lambda_1\mathcal{L}_{\text{imp}}+\lambda_2\mathcal{L}_{\text{cls}}$

	%\item Effect of $\mathbf{W}_{\boldsymbol{\alpha}}$ to be the diagonal or full weight matrix: For the diagonal $\mathbf{W}_{\boldsymbol{\alpha}}$ case, each variable's $\mathbf{u}_t$ is independently modeled and for full weight matrix case, an interaction of $\mathbf{u}_t$ between the variables is modeled.
\end{itemize}

%%%%%%%%%%%%%%%%%%%%%%%%%%%%%%%%%%%%%%%%%%%%%%
\section{Discussion}
\label{sec:Discussion}

% \begin{figure}[h!]
%     \centering{\includegraphics[width=\columnwidth]{./figures/UNC_MAE.pdf}}
%     \caption{Visualization of uncertainty values depending on MAE values.}\label{fig:MAE_UNC_Corr}
% \end{figure}

%%%%%%%%%%%%%%%%%%%%%%%%%%%%%%%%%%%%%%%%%%%%%%
\subsection{Visualization of Imputation Estimates}
\label{subsec:Analysis of Imputation Estimates}
Fig. \ref{fig:imputation_overall} compares actual observations and our model predictions with their uncertainties on the PhysioNet dataset under both 5\% and 10\% masking scenarios. Our model tends to produce temporally smooth curves and also exhibits different levels of uncertainty over the time-series. It is noteworthy that the uncertainty estimates correlate qualitatively with the missingness of the features and the noise levels of the observations. This helps clinicians make informed decisions regarding the fidelity they should have in the model.

% \begin{itemize}
%     \item uncertainty correlate with missingness (즉, temporally missing이 심할 수록 uncertainty가 높은 경향)
%     \item uncertainty가 높을 때/낮을 때 각각 MAE값이 어떤지 비교
%     \item comparison per scenario
% \end{itemize}

% as is depicted in Figure 

% Visualization of Ground truth, VRNN predicted dec mean, VRNN predicted dec std (uncertainties)

%%%%%%%%%%%%%%%%%%%%%%%%%%%%%%%%%%%%%%%%%%%%%%
\subsection{Analysis of Model Behavior}
\label{subsec:Analysis of Model Behavior}
{\chg 
In the results of the missing value imputation described in Section \ref{subsec:Result_of_Missing_Value_Imputation}, our proposed method showed slightly higher MAE scores. However, it should be noted that the ultimate goal of our study is to correctly predict the mortality by leveraging a missing value imputation. Further, owing to the noisy observations in practice, we explicated the uncertain noisy factors in imputing the missing values and exploited such uncertain factors into our prediction model. Thus, whereas the MAE score of our method is higher than that of the competing methods, by better reflecting the noisy factors or imputation values in terms of uncertainty, we could achieve a better mortality prediction performance, which is imperative in a clinical setting.

Hence, we further investigated the behavior of our model to represent the uncertainties for the imputation estimates. For this, we calculated the Pearson's correlation between the MAE and the uncertainty values using a testset of the PhysioNet dataset under both 5\% and 10\% masking scenarios. For each data instance, we obtained the MAE between the ground truth and predicted imputations with the corresponding uncertainties.

Fig. \ref{fig:imputation_overall} shows a scatter plot of the uncertainty values depending on the MAE values under 5\% and 10\% masking scenarios, respectively. For the $5\%$ masking scenario, the average correlation coefficient is 0.406 with $p=1.299e^{-7}$  for null hypothesis that there is no correlation between MAE and uncertainty, and for the $10\%$ masking scenario, the correlation coefficient is 0.338 with $p=1.406e^{-5}$. 
These experimental results indicate that our model can provide sufficient information regarding the fidelity of the model by largely predicting the uncertainty, depending on the MAE value, even if the estimated imputations are far from the actual observation.

% 5\%
% Total average correlation: 0.40624619478769997
% Total average p-value : 1.2987033978762547e-07

% 10\%
% Total average correlation: 0.3389963372854419
% Total average p-value : 1.405688777287393e-05

% {\todo LRP analysis}

% future works
% its predictions, which reflects the estimated uncertainty into mortality predictions. 

}

%%%%%%%%%%%%%%%%%%%%%%%%%%%%%%%%%%%%%%%%%%%%%%
% \subsection{t-SNE Visualization of Latent Space}
% \label{subsec:t-SNE Visualization of Latent Space}

%%%%%%%%%%%%%%%%%%%%%%%%%%%%%%%%%%%%%%%%%%%%%%
% \subsection{Model Weight Analysis}
% \label{subsec:Model Weight Analysis}

%%%%%%%%%%%%%%%%%%%%%%%%%%%%%%%%%%%%%%%%%%%%%%
\section{Conclusion}
In this work, we proposed a novel uncertainty-gated stochastic sequential imputation method that extends the VRNN for mortality prediction with EHR data. By leveraging VRNN as our base model, we handle representation learning, missing value imputation, and in-hospital mortality prediction in a single stream. In addition, we proposed the novel GRU cell, in which temporal information encoding in hidden states is updated by propagating the uncertainty for the inferred distribution over the variables. We validated the effectiveness of our method over the public MIMIC-III and PhysioNet challenge 2012 datasets by comparing with and outperforming to the state-of-the-art methods considered in our experiments for mortality prediction. Furthermore, we identified the behavior of the model that well represented the uncertainties for the imputed estimates, which indicated a high correlation between the calculated MAE and the uncertainty.

\appendices

%\section{The Other Activation Patterns}

\section*{Acknowledgment}
This work was supported by Institute of Information \& communications Technology Planning \& Evaluation (IITP) grant funded by the Korea government (MSIT) (No. 2017-0-00053, A technology development of artificial intelligence doctors for cardiovascular disease, and No. 2019-0-00079, Department of Artificial Intelligence (Korea University)).

% {\todo This work was supported by Institute for Information \& Communications Technology Promotion (IITP) grant funded by the Korea government (No. 2017-0-00451, Development of BCI based Brain and Cognitive Computing Technology for Recognizing User's Intentions using Deep Learning).}
% This work was supported by Institute of Information & communications Technology Planning & Evaluation (IITP) grant funded by the Korea government (MSIT) (No. 2019-0-00079, Department of Artificial Intelligence(Korea University))

\section*{Data Availability}
MIMIC-III database analyzed in this study is available on PhysioNet repository. %The code to generate the datasets used in this study is available on 

\ifCLASSOPTIONcaptionsoff
  \newpage
\fi

\bibliographystyle{IEEEtran}
\bibliography{arXiv}

% \begin{IEEEbiography}[{\includegraphics[width=1in,height=1.25in,clip,keepaspectratio]{mshell}}]{Eunji Jun}
% 	received the B.S. degree in electronics engineering from Ewha Womans University, Seoul, South Korea, in 2017. She is currently pursuing the Ph.D. degree with the Department of Brain Cognitive Engineering, Korea University, Seoul, South Korea. 
    
%     Her current research interests include biomedical machine learning, medical image analysis, and healthcare. 
% \end{IEEEbiography}

% \begin{IEEEbiography}[{\includegraphics[width=1in,height=1.25in,clip,keepaspectratio]{mshell}}]{Ahmad Mulyadi Wisnu}
% \end{IEEEbiography}

% \begin{IEEEbiography}[{\includegraphics[width=1in,height=1.25in,clip,keepaspectratio]{mshell}}]{Heung-Il Suk}
%   received the Ph.D. degree in computer science and engineering from Korea University, Seoul, South Korea, in 2012. 
   
%   From 2012 to 2014, he was a Post-Doctoral Research Associate with the University of North Carolina at Chapel Hill, Chapel Hill, NC, USA. He is currently an Associate Professor with the Department of Brain and Cognitive Engineering and the Department of Artificial Intelligence, Korea University. His current research interests include machine learning, pattern recognition, biomedical image analysis, brain-computer interface, and healthcare.
   
%   Dr.Suk serving as a Program Committee or Reviewer for NeurIPS, ICML, AISTATS, ICLR, AAAI, MICCAI, and so on.
% \end{IEEEbiography}

\end{document}